\documentclass{article}
\usepackage[english]{babel}
\usepackage{titlesec}

\usepackage[letterpaper,top=2cm,bottom=2cm,left=3cm,right=3cm,marginparwidth=1.75cm]{geometry}

\usepackage{amsmath}
\usepackage{graphicx}   % 그림 삽입
\usepackage[colorlinks=true, allcolors=blue]{hyperref}
\usepackage[numbers,sort&compress]{natbib}
\usepackage{kotex}
\usepackage{listings}
\usepackage{booktabs}
\usepackage{multirow}
\usepackage{caption}    % 캡션 커스터마이즈
\usepackage{subcaption} % 서브피겨가 필요할 때
\usepackage{float}
\usepackage{threeparttable}
\usepackage{array}
\newcolumntype{T}[1]{>{\ttfamily\raggedright\arraybackslash}p{#1}}

\usepackage{pgfplots}
\pgfplotsset{compat=1.18}

\title{SDS KoPub VDR: A Benchmark Dataset for Visual Document Retrieval in Korean Public Documents}

\author{%
  Jaehoon Lee, Sohyun Kim, Wanggeun Park\\[0.3em]
  Geon Lee, Seungkyung Kim, Minyoung Lee\\[0.5em]
  Samsung SDS, AI Advanced Research Lab\\[0.5em]
  \texttt{jhlee19.lee@samsung.com, sh\_sds.kim@samsung.com, wking.park@samsung.com}\\
  \texttt{go.lee@samsung.com, seungkyung.kim@samsung.com, miny.lee@samsung.com}
}
\date{}  
\begin{document}
\maketitle

\begin{abstract}
Existing benchmarks for visual document retrieval (VDR) largely overlook non-English languages and the structural complexity of official publications. To address this gap, we introduce SDS KoPub VDR, the first large-scale, public benchmark for retrieving and understanding Korean public documents. The benchmark is built upon 361 real-world documents, including 256 files under the KOGL Type 1 license and 105 from official legal portals, capturing complex visual elements like tables, charts, and multi-column layouts. To establish a reliable evaluation set, we constructed 600 query-page-answer triples. These were initially generated using multimodal models (e.g., GPT-4o) and subsequently underwent human verification to ensure factual accuracy and contextual relevance. The queries span six major public domains and are categorized by the reasoning modality required: text-based, visual-based, and cross-modal. We evaluate SDS KoPub VDR on two complementary tasks: (1) text-only retrieval and (2) multimodal retrieval, which leverages visual features alongside text. This dual-task evaluation reveals substantial performance gaps, particularly in multimodal scenarios requiring cross-modal reasoning, even for state-of-the-art models. As a foundational resource, SDS KoPub VDR enables rigorous and fine-grained evaluation and provides a roadmap for advancing multimodal AI in real-world document intelligence. The dataset is publicly available at https://huggingface.co/datasets/SamsungSDS-Research/SDS-KoPub-VDR-Benchmark    
\end{abstract}

\section{Introduction}

With the rapid advancement of large language models (LLMs), retrieval-augmented generation (RAG) has become a powerful paradigm for combining external knowledge with generative models\cite{lewis2021retrievalaugmentedgenerationknowledgeintensivenlp}. However, the overall performance of RAG systems remains critically dependent on the accuracy of the retrieval stage. In particular, accurately retrieving relevant information from complex visual documents — documents containing tables, charts, figures, and multi-column layouts — remains one of the most significant unresolved challenges. Many government reports, public white papers, and statistical yearbooks encode key information not only in text but also in visual cues, making VDR a fundamental prerequisite for high-performing multimodal RAG systems. Retrieval errors in such settings often cascade into system-level failures, causing models to hallucinate or produce evasive answers, as noted in several studies \cite{gao2024retrievalaugmentedgenerationlargelanguage}. Consequently, developing reliable methods and benchmarks for evaluating retrieval performance in multimodal contexts has become increasingly crucial.

Despite its importance, existing benchmarks have notable limitations in evaluating VDR as an independent capability. Prominent QA benchmarks such as SQuAD \cite{rajpurkar2016squad100000questionsmachine} focus solely on textual data, while document-based Visual Question Answer(VQA) datasets like DocVQA \cite{mathew2021docvqadatasetvqadocument} and InfographicVQA \cite{mathew2021infographicvqa} primarily measure the final answer generation performance. These benchmarks neither isolate the impact of retrieval errors on downstream tasks nor capture the structural and linguistic complexity of Korean public documents, which often feature multi-column layouts, dense tables, and visual cues critical to information understanding. Moreover, the lack of a standardized testbed for systematically comparing and improving multimodal retrievers remains a significant barrier to progress in this field.

To address these limitations, we introduce SDS KoPub VDR Benchmark, the first large-scale, publicly available benchmark for VDR in Korean public documents. The dataset comprises 361 real-world documents (40,781 pages), including 256 PDFs under the KOGL Type 1 public license and 105 collected from official legal portals. It also includes 600 query–page–answer triples, generated with multimodal language models (GPT-4o, Qwen2.5-VL-72B) and refined through rigorous human verification. Queries span six major public domains — society, environment, education, industry, diplomacy, and finance — and are systematically categorized into three types based on the reasoning modality required: text-based, visual-based, and cross-modal.

Beyond proposing a new dataset, our study aims to answer two fundamental research questions:
\begin{itemize}
    \item \textbf{RQ1:} To what extent does incorporating visual information (e.g., page images) improve retrieval performance compared to traditional text-only approaches?
    \item \textbf{RQ2:} How effectively do current multimodal models handle complex visual reasoning tasks --- such as identifying specific table values or interpreting chart trends --- and which query types remain particularly challenging?
\end{itemize}

To investigate these questions, we design two complementary evaluation tasks:
\begin{itemize}
    \item \textbf{Task 1: Text-only Retrieval} -- Retrieval is performed using text embeddings extracted from the outputs of PDF parsing tools (e.g., PyPDF), enabling a quantitative comparison of performance gaps between text-only and multimodal approaches.
    \item \textbf{Task 2: Multimodal Retrieval} -- Retrieval leverages multimodal embeddings that incorporate both textual and visual information, enabling evaluation of the contribution of visual cues and the difficulty of cross-modal reasoning.
\end{itemize}

Our baseline experiments reveal that the multimodal approach provides clear advantages over text-only retrieval, particularly for visual and cross-modal queries where information is encoded in non-textual elements. Text-based methods often fail to capture these multimodal cues, whereas multimodal retrieval leverages both textual and visual representations to produce more contextually grounded results. Nonetheless, its ability to handle complex queries remains constrained, underscoring the importance of developing more advanced visual reasoning capabilities for future multimodal retrievers. The SDS KoPub VDR benchmark is specifically designed to capture these challenges with precision and to serve as a foundation for advancing multimodal retrieval architectures.

The main contributions of this paper are summarized as follows:
\begin{itemize}
    \item \textbf{A Large-Scale, Realistic Benchmark:} We release the first large-scale VDR dataset for Korean public documents, comprising 361 documents (40,781 pages) and 600 queries across six domains and three query types, faithfully reflecting real-world complexity.
    \item \textbf{Task Design Grounded in Key Research Questions:} By separating text-only and multimodal retrieval tasks, the benchmark provides a principled framework for analyzing the contribution of visual information (RQ1) and the difficulty of different query types (RQ2).
    \item \textbf{Reproducible Evaluation Protocol and Baselines:} We provide an evaluation protocol based on Recall@k and nDCG@k and report baseline performance, enabling fair and meaningful comparison across future studies.
    \item \textbf{Data Governance and Reliability:} We ensure data transparency by clearly documenting data sources, licensing, and the full pipeline from collection to validation, thereby guaranteeing reproducibility and trustworthiness.
\end{itemize}

Overall, SDS KoPub VDR Benchmark is designed not merely to expand dataset scale but to serve as a standardized testbed that captures the real-world challenges inherent in complex Korean public documents. Beyond simple performance comparison, it provides a precise analytical tool for diagnosing multimodal models’ visual reasoning failures and guiding the development of more robust RAG systems. We expect this benchmark to set a new standard for multimodal document understanding research and to accelerate progress in this rapidly evolving field.

\section{Related Work}

\subsection{RAG}
RAG combines large language models with retrieval systems to ground generated outputs in external knowledge.
Early works such as DPR\cite{karpukhin2020densepassageretrievalopendomain}, REALM\cite{guu2020realmretrievalaugmentedlanguagemodel}, and RAG established dense text retrieval frameworks, while later models (e.g., Atlas\cite{izacard2022atlasfewshotlearningretrieval}, FiD\cite{izacard2021leveragingpassageretrievalgenerative}) improved context aggregation and efficiency.
Recent studies including CoT-RAG\cite{izacard2022atlasfewshotlearningretrieval} and Self-RAG\cite{asai2023selfraglearningretrievegenerate} extend RAG toward reasoning-guided and adaptive retrieval, showing that retrieval selection can be learned jointly with generation. However, these methods focus primarily on textual retrieval, overlooking visually structured information such as tables, diagrams, and layouts that are common in real-world documents. This limitation motivates VDR, where models must locate and interpret visual evidence rather than purely textual passages.

\subsection{Visual Document Understanding (VDU) \& VQA}
VDU and VQA explore the integration of text, layout, and image modalities to interpret complex document pages.
Benchmarks like DocVQA\cite{mathew2021docvqadatasetvqadocument} and InfographicVQA\cite{mathew2021infographicvqa} examine document-level reasoning, while ChartQA\cite{masry2022chartqabenchmarkquestionanswering} targets numerical reasoning over graphical plots.
Layout-aware architectures (e.g., LayoutLMv3\cite{huang2022layoutlmv3pretrainingdocumentai}, DocLayNet\cite{Pfitzmann_2022}, Pix2Struct\cite{lee2023pix2structscreenshotparsingpretraining}) further enhance structural understanding by capturing layout-text dependencies.
Yet, these efforts primarily assess in-page understanding—how well a model answers a question given a single document image rather than retrieval across large collections.
In contrast, VDR requires identifying relevant evidence pages from large repositories, demanding both semantic and spatial reasoning across multimodal inputs.

\subsection{Multimdoal Embedding Model}
The rise of vision-language embedding models such as CLIP\cite{radford2021learningtransferablevisualmodels} and BLIP\cite{li2022blipbootstrappinglanguageimagepretraining} demonstrated the potential of unified text–image spaces.
Later works like VLM2Vec\cite{jiang2025vlm2vectrainingvisionlanguagemodels}, E5-V\cite{jiang2024e5vuniversalembeddingsmultimodal}, and Nomic-Embed-Multimodal\cite{nomicembedmultimodal2025} extended this idea to structured and document-level data, incorporating layout and visual components into representation learning.
However, these models are trained mostly on web-scale English datasets and lack domain or language specialization.
This results in limited robustness when applied to structured public-sector documents that feature mixed text and visual data.
Thus, SDS KoPub VDR addresses the need for domain-specific, Korean-language multimodal embeddings that handle heterogeneous document layouts.

\subsection{VDR benchmark}
Recent benchmarks broaden multimodal retrieval evaluation. ViDoRe\cite{faysse2025colpaliefficientdocumentretrieval} introduces multi-hop visual reasoning, MMDocIR\cite{dong2025mmdocirbenchmarkingmultimodalretrieval} focuses on long-document retrieval, and VisR-Bench\cite{chen2025visrbenchempiricalstudyvisual} evaluates multilingual document search.
Despite such progress, most datasets remain English-centric, rely on web or synthetic images, and rarely provide open licensing or domain diversity.
Korean resources such as RAG-Evaluation-Dataset-KO\cite{allganize_rag_eval_ko} assess text-only retrieval and lack visual structure.
Our proposed SDS KoPub VDR fills this gap as the first large-scale, page-level benchmark for Korean public documents, integrating text, layout, and visual reasoning under an open license to support multimodal RAG research.

\section{Dataset Construction}
\begin{figure}[htb]
    \centering
    \includegraphics[width=0.9\textwidth]{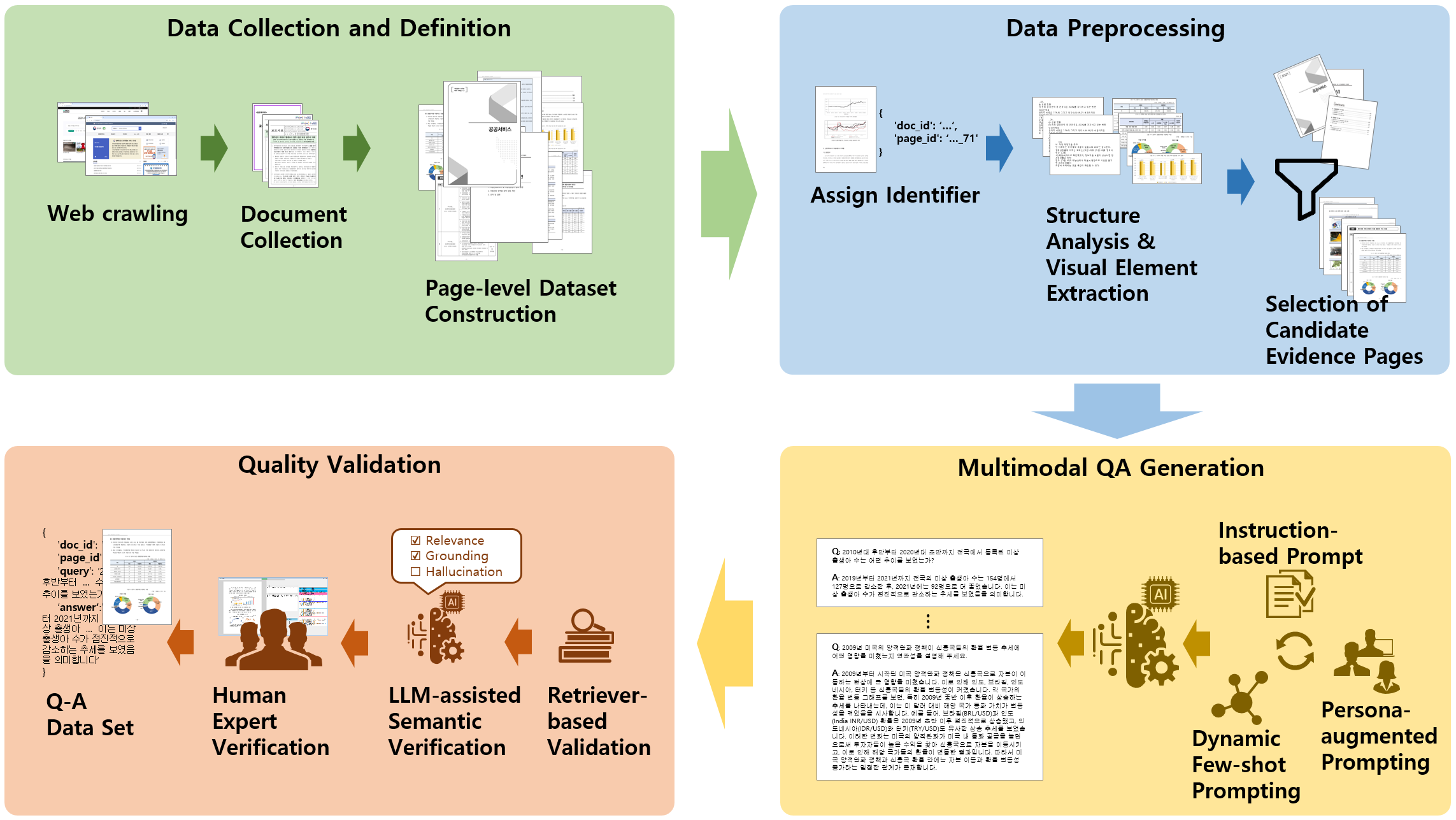}
    \caption{SDS KoPub VDR Benchmark Dataset Construction Process}
    \label{fig:data-flow}
\end{figure}

This section describes the process of constructing a high-quality multimodal QA benchmark based on Korean public documents.
The overall data construction flow, consisting of four main stages—Data Collection and Definition, Data Preprocessing, Multimodal QA Generation, and Quality Validation—is illustrated in Figure \ref{fig:data-flow}.

\subsection{Data Collection and Definition}
\subsubsection{Source Document Collection}
The data collection primarily targeted two sources: (1) Administrative materials from national and local governments, freely available under the Korea Open Government License (KOGL) Type 1. (2) Legal documents, including statutes, public notices, and official guidelines, provided by the Korea Ministry of Government Legislation.

All collected documents satisfy the conditions for creating and redistributing derivative works and were gathered in compliance with public data reuse guidelines. From an initial pool of 18,901 documents, we curated a select set of 361 documents based on the inclusion of unstructured visual elements (e.g., tables, charts, diagrams), content diversity, and domain representativeness. This final collection consists of policy reports, statistical yearbooks, implementation plans, guidelines, and legal compendiums, spanning six domains: society, environment, education, industry, diplomacy, and finance. This composition enables a comprehensive evaluation of a model's capabilities, including domain generalization and the understanding of unstructured information, thereby addressing the limitations of existing text-centric benchmarks.

\subsubsection{Page-level Dataset Construction}
Public administrative documents are typically provided as PDFs ranging from tens to hundreds of pages, with each page containing distinct visual and structural information. Recognizing that the input unit for multimodal retrieval models is often a `page' or `segment' rather than the `entire document', we established the page as the fundamental unit for our benchmark construction. By segmenting the 361 selected documents, we obtained a total of 40,781 pages. From this collection, we designated information-rich pages as the ``Ground Truth Evidence" for QA generation, while incorporating all pages from the source document into the searchable ``Corpus". This design emulates a realistic retrieval scenario, allowing for a precise evaluation of a model's ability to identify the single most relevant evidence page.

\subsubsection{Query Type Definition}
Our benchmark is designed to assess not just simple text retrieval but also visual information understanding and complex reasoning abilities. To this end, we define three query types based on the modality of the evidence source required to derive the answer, as outlined in Table \ref{tbl:query_type_definitions}.

\begin{table}[htb]
\centering
\caption{Query Type Definitions}
\label{tbl:query_type_definitions}
\begin{tabular}{p{2.5cm} p{11cm}}
\toprule
\textbf{Query Type} & \textbf{Definition} \\
\midrule
\textbf{Text} & Queries that can be answered using only the body text extracted from documents through OCR or text-based PDF parsing tools such as PyPDF. \\
\textbf{Visual} & Queries that require information exclusively from visual elements such as tables, graphs, or diagrams.  \\
\textbf{Cross} & Queries that necessitate referencing both textual and visual elements to formulate a complete answer. \\
\bottomrule
\end{tabular}
\end{table}

\begin{figure}[htbp]
    \centering
    \includegraphics[width=0.9\textwidth]{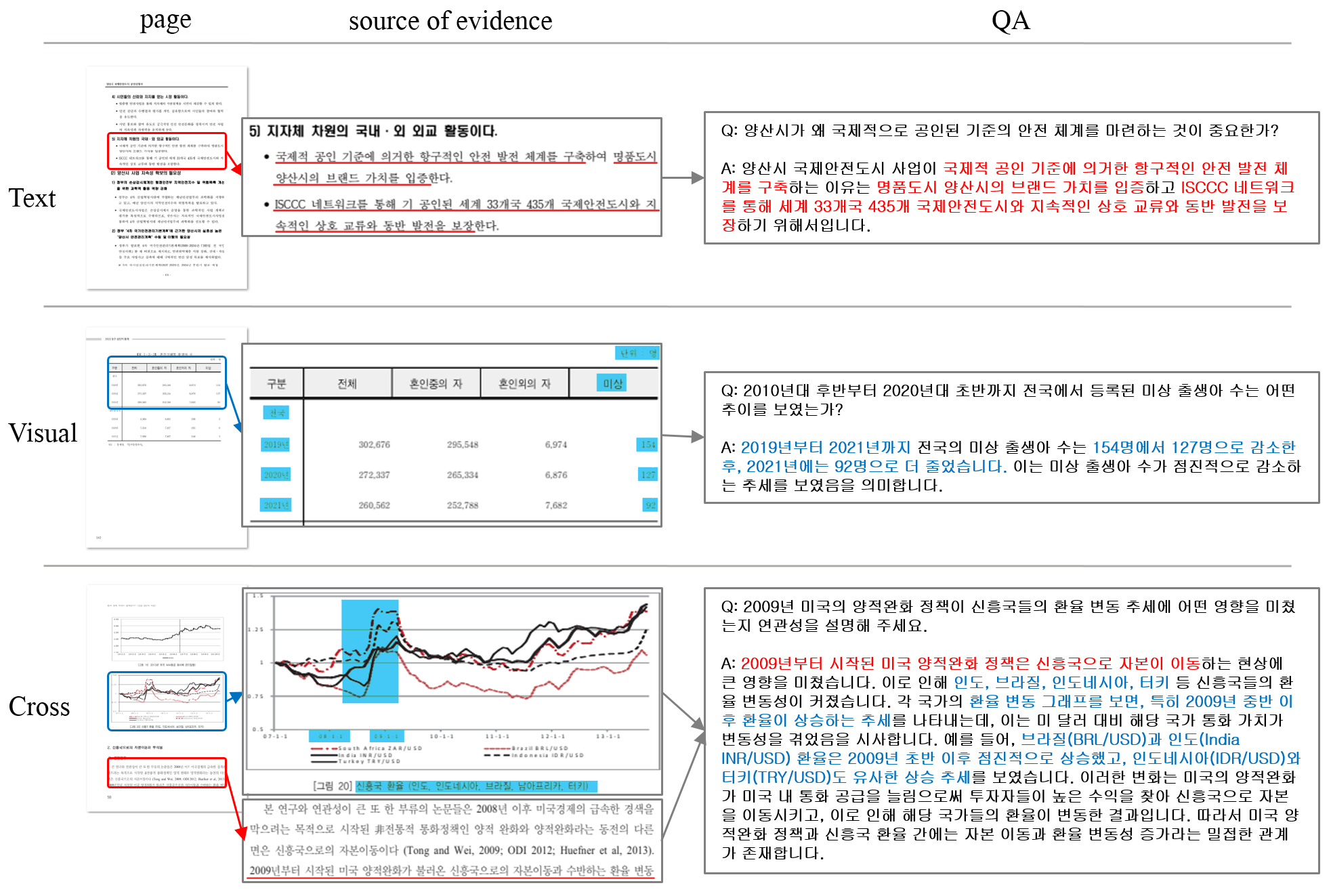}
    \caption{Query type example}
    \label{fig:query_type_definitions}
\end{figure}

This classification is based on the source of the evidence, not the linguistic form of the query. After analyzing each page to identify the location of the answer's evidence, we categorize the query as Text, Visual, or Cross according to the evidence modality. We strategically increased the proportion of Visual and Cross queries to enhance the benchmark's discriminative power in areas where performance variance among models is most pronounced. Examples for each query type are presented in Figure \ref{fig:query_type_definitions}.

\subsection{Data Preprocessing Pipeline}
To transform the collected PDF documents into structured data suitable for multimodal QA generation and retrieval evaluation, we designed a three-stage preprocessing pipeline. This pipeline standardizes document formats, extracts textual and visual information into a refined structure, and selects semantically rich pages to serve as evidence units.

\subsubsection{Page Conversion and Identifier Assignment}
All source documents were split into individual pages and converted into PNG images at a 300 DPI resolution. Each page was assigned a unique identifier (doc\_id, page\_id) to ensure consistent referencing across the entire dataset. We also extracted structural metadata, such as the total page count, current page number, section titles, and publication information, for use in subsequent QA generation and evaluation phases. This ensures that all following procedures operate on a unified, page-centric data representation.

\subsubsection{Document Layout Analysis and Visual Element Extraction}
Korean public documents often feature complex layouts, including multi-column text, tables, charts, and diagrams, which can lead to significant information loss if processed with simple OCR alone. To address this, we employed Docling's \cite{Docling} Advanced PDF Understanding module to analyze the logical structure of each document (e.g., titles, paragraphs, subsections) and independently identify non-textual visual elements. Each visual element was cropped and saved at an 72 DPI resolution. This approach preserves essential information while minimizing image size by removing unnecessary margins, enabling both clear information delivery and efficient token usage during data generation. This process yielded a data representation where textual and visual information are explicitly segregated, contributing to improved modality alignment and evidence-based grounding for the QA generation models.

\subsubsection{Selection of Candidate Evidence Pages}
From the set of all processed pages, we selected candidates for QA generation that met all of the following criteria: 
\begin{itemize}
\item Contains at least 300 characters of text. 
\item Includes one or more non-textual visual elements (table, chart, diagram). 
\item Excludes non-informative pages such as covers and tables of contents.
\end{itemize}

Pages satisfying these criteria constitute the ``Evidence Candidate Pool" which comprises high-quality data with a balanced representation of textual and visual information. This refined, page-level data forms the foundation for evaluating multimodal QA and retrieval models in a realistic information-seeking context.

\subsection{Multimodal Query–Answer Generation}
Leveraging the preprocessed page data and metadata, we automatically generated QA pairs using Multimodal LLMs, including GPT-4o and Qwen2.5-VL-72B. This stage was designed to replicate realistic search query-response scenarios rather than simple descriptions, combining the following three prompt engineering strategies.

\subsubsection{Instruction-based Prompting}
We instructed the models to generate QA pairs under the premise that a user is asking a question without having seen the page. The instructions strictly mandated that answers must be derived only from the evidence present on the given page. The model first determines the suitability of the page for QA generation and proceeds only if the page is information-rich. Question types were diversified to include factual, definitional, relational, and causal questions, and were designed to incorporate non-textual evidence to prevent monotonic outputs. Answers were required to be clear and complete sentences to serve as ground truth for evaluation.

\subsubsection{Persona-augmented Prompting}
To encourage the generation of queries that reflect realistic contexts and user intent, we assigned domain-specific personas to the model, such as `Policy Maker', `Citizen Petitioner', and `Journalist'. This approach moves beyond simple summarization to incorporate contextual intent, prompting a reasoning process that involves information seeking, comprehension, and synthesis. This resulted in the generation of more complex and challenging QA pairs that closely mirror real-world user behavior.

\subsubsection{Dynamic Few-shot Prompting}
During the initial document crawling phase, we collected Q\&A, FAQ, and case study documents from various public institutions to build a domain-specific ``Few-shot Pool". When generating a prompt for a specific page, our system dynamically selects relevant examples from this pool by performing a keyword-based search using the page's metadata (e.g., source, publishing agency, domain, title). This differs from static few-shot prompting in that it actively selects reference examples tailored to the topic and domain of each page. This method enabled the generation of QA pairs that reflect domain-specific stylistic conventions and response formats, closely aligning with those of actual public inquiries.

\subsection{Quality Validation}
To ensure the reliability and consistency of the generated QA dataset, we implemented a three-stage quality validation process: (1) Retriever-based Validation, (2) LLM-assisted Semantic Verification, and (3) Human Expert Verification.

\subsubsection{Retriever-based Validation}
In our benchmark, each query is mapped to a single, most appropriate evidence page as its ground truth. Therefore, it is crucial to verify that there are no duplicate answer pages or evidence conflicts within the dataset. We applied a BM25-based retriever to the page texts and QA pairs to identify duplicate pages and ensure that partially overlapping evidence paragraphs were not repeated at the token level. Furthermore, to confirm that our queries evaluate contextual understanding rather than simple keyword matching, we validated the difficulty by applying query rewriting techniques (e.g., synonym substitution, paraphrasing) to queries with high retrieval scores.

\subsubsection{LLM-assisted Semantic Verification}
We performed automated semantic verification on all QA pairs using GPT-4.5. The model conducted a list-wise scoring comparison across the top-retrieved results, evaluating the following three criteria: 

\begin{itemize}
\item Context Relevance: Can the query be adequately answered based on the elements within the page?  
\item Answer Grounding / Faithfulness: Does the answer correspond precisely with the content of the page? 
\item Hallucination Check: Does the answer contain external knowledge or fabricated information?
\end{itemize}

QA pairs with scores below a predefined threshold or instances where the ground truth page was not ranked highest were flagged as low-quality and excluded from the dataset.

\subsubsection{Human Expert Verification}
Finally, QA pairs that passed the automated validation stages underwent an exhaustive manual review by our researchers using a dedicated annotation tool. Each pair was compared directly against the source page. The verification checklist included:

\begin{itemize}
\item Query Clarity: Is the query specific, unambiguous, and not open to multiple interpretations?  
\item Answer Correctness: Does the answer accurately reflect the information on the page? 
\item Evidence Alignment: Does the ground truth field correctly reference the location of the evidence? 
\item Type Appropriateness: Does the query accurately conform to its declared modality (text, visual, cross)?
\end{itemize}

This process corrected subtle errors missed during automated verification and manually rectified cases of evidence mismatch and ambiguous phrasing. The final QA set that passed this exhaustive human review constitutes a high-quality multimodal QA benchmark, satisfying both the quantitative reliability of automated checks and the qualitative standards of human expert judgment.

\section{Dataset Statistics and Analysis}
This section presents a quantitative analysis of the SDS KoPub VDR benchmark, detailing the composition of its documents, pages, and multimodal elements, and discussing the implications of its structural characteristics for real-world evaluation and research applications.

\subsection{Document and Page Distribution}
The benchmark comprises a diverse collection of documents including policy reports, legal commentaries, and project plans. Document length varies widely—from several dozen to several hundred pages—reflecting the structural complexity and heterogeneity of real Korean public documents. Table \ref{tbl:source-dist} summarizes the distribution of documents by source institution and content type.

\begin{table}[htb]
\centering
\begin{threeparttable}
\caption{Document and Page Distribution}
\label{tbl:source-dist}
\begin{tabular}{l l r r r}
\toprule
Source & Topic & \# Docs & \# Pages & Avg. Words/Page \\
\midrule
NAS\tnote{1} 
  & Reports on diplomatic trends, international affairs & 7 & 366 & 215.45 \\
NARS\tnote{2} 
  & Reports on administrative actions, legislative cases & 125 & 8{,}176 & 180.22 \\
NABO\tnote{3} 
  & Fiscal analyses, project evaluation reports & 2 & 310 & 278.41 \\
PRISM\tnote{4} 
  & Research on social, environmental, and industrial policy & 122 & 31{,}500 & 244.23 \\
MOLEG\tnote{5} 
  & Legal guides, statutory interpretations, case studies & 105 & 429 & 218.69 \\
\bottomrule
\end{tabular}
\begin{tablenotes}[flushleft]
\footnotesize
\item[1] National Assembly Secretariat.
\item[2] National Assembly Research Service.
\item[3] National Assembly Budget Office.
\item[4] Policy Research Information Service \& Management.
\item[5] Ministry of Government Legislation.
\end{tablenotes}
\end{threeparttable}
\end{table}

\begin{figure}[htb]
    \centering
    \includegraphics[width=1\textwidth]{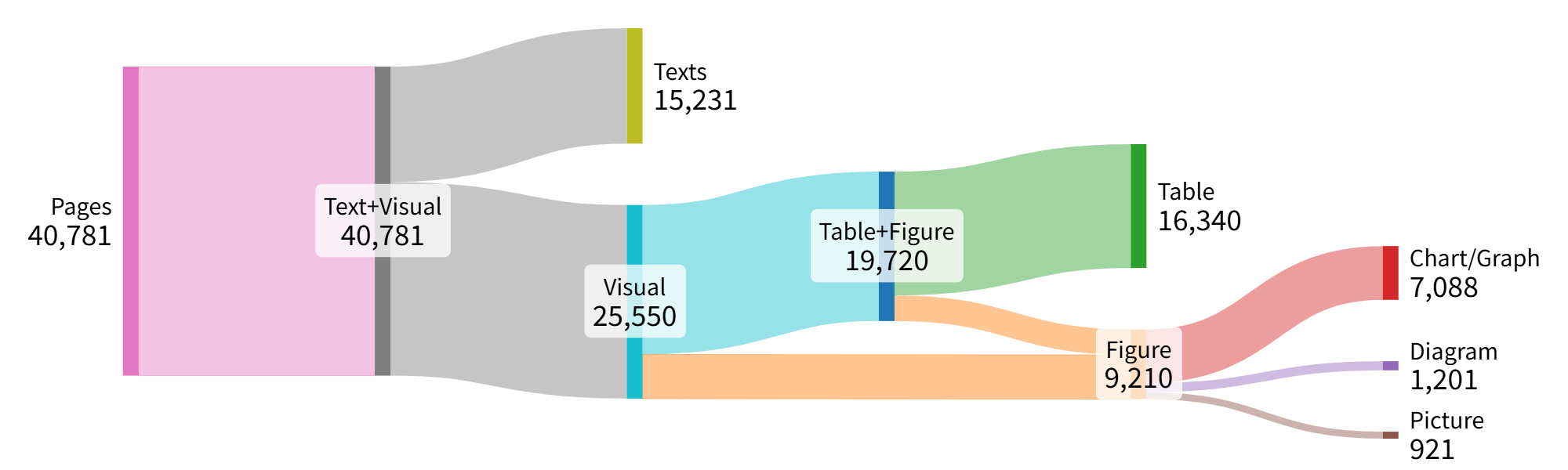}
    \caption{Distribution of page compositions and visual elements}
    \label{fig:dist-page-comp}
\end{figure}

A key characteristic of SDS KoPub VDR lies in its focus on pages containing structured and visual information such as tables, charts, graphs, and diagrams, rather than text-only content. Figure \ref{fig:dist-page-comp} illustrates the overall distribution of visual elements across all pages.

An analysis of the page-level composition reveals that out of 40,781 total pages, 15,231 (37.3\%) are purely text-based, while the remaining 25,550 pages (62.7\%) contain one or more visual elements. These visually-rich pages are categorized into two main groups. 

First, the Table+Figure group (19,720 pages) contains tables, where 16,340 pages (40.1\% of total) are table-centric, and 3,380 pages (8.3\%) feature a composite structure with both tables and other figures. Second, the Figure-only group (9,210 pages) consists of pages centered on visual data, including 7,088 pages (17.4\%) with charts or graphs, 1,201 pages (2.9\%) with diagrams, and 921 pages (2.3\%) with pictures. This distribution highlights a key structural characteristic of Korean public documents: the active use of visual representations to reinforce logical and quantitative information. In particular, the Table+Figure category creates complex contexts where models must perform cross-modal reasoning—correlating information between text, tables, and figures—to derive correct answers. 

These pages present a significant challenge for multimodal RAG and document understanding models, demanding a level of visual compositionality that is difficult for single-modality systems to address.

\subsection{Query-Answer Analysis}
The finalized benchmark comprises 600 QA pairs uniformly distributed across six major domains—society, environment, education, industry, diplomacy, and finance. The distribution of query types within each domain is shown in Table \ref{tbl:qa-type-by-domain}.

\begin{table}[htp]
\centering
\caption{Distribution of query types across domains}
\label{tbl:qa-type-by-domain}
\begin{tabular}{lrrrr}
\toprule
Domain & Cross & Visual & Text & Total \\
\midrule
Education  & 61 & 29 & 10 & 100 \\
Finance    & 54 & 26 & 20 & 100 \\
Society    & 57 & 29 & 14 & 100 \\
Industry   & 76 & 13 & 11 & 100 \\
Diplomacy  & 54 & 26 & 20 & 100 \\
Environment& 34 & 38 & 28 & 100 \\
\midrule
\textbf{Total} & \textbf{336} & \textbf{161} & \textbf{103} & \textbf{600} \\
\bottomrule
\end{tabular}
\end{table}

By query type, Visual (161) and Cross (336) queries constitute 82.8\% of the entire set. This deliberately skewed distribution reflects the benchmark's primary objective: to move beyond simple text retrieval and rigorously evaluate visual understanding and cross-modal reasoning capabilities. Furthermore, the balanced distribution of 100 QA pairs per domain ensures a fair and robust assessment of a model's domain generalization performance.

A closer look at the domain-specific distributions reveals that document characteristics directly influence query composition. For example, the industry domain, which is rich in documents with numerical data and performance reports, has a higher proportion of queries requiring cross-modal reasoning. Conversely, the environment domain frequently utilizes visual aids like maps, charts, and illustrated guidelines, leading to a relatively higher share of Visual type queries. This variance suggests that models must employ adaptive retrieval and reasoning strategies tailored to the unique characteristics of each domain. This provides a foundation for a fine-grained analysis of a model's strengths and weaknesses across diverse document types.

\section{Experimental Setup and Results}
This section presents the retrieval evaluation tasks defined in the SDS KoPub VDR benchmark, the corresponding evaluation metrics and baseline models, and the overall experimental results. The benchmark is designed to quantitatively measure a model’s ability to leverage both textual semantics and visual document structures for IR within Korean public-sector documents.

\subsection{Retrieval Tasks}
\subsubsection{Task 1: Text-only Retrieval}
The first task establishes a traditional IR baseline, where both queries and documents are represented solely by their pdf-extracted(e.g., PyPDF) textual content.
This task serves as a baseline for comparison with multimodal approaches and quantifies the impact of the quality of the text parser on downstream retrieval accuracy.
An input query (e.g., “What are the procedures for environmental impact assessment?”) is encoded as a textual embedding, while each document page is represented by an embedding derived from its transcribed content.
The objective is to retrieve the page that best matches the query’s semantics.
However, this approach is inherently constrained by page parsing degradation on pages containing complex visual elements such as tables or charts.

\subsubsection{Task 2: Multimodal Retrieval}
The second task extends the evaluation to multimodal document understanding by jointly leveraging visual and textual information.
It aims to evaluate a model’s ability to utilize layout-sensitive structural signals (e.g. tables, graphs, multicolumn formats) and achieve semantic alignment between textual and visual representations.
Each document page is processed as a full-page image, and a joint multimodal embedding is constructed by combining its visual features with PDF-parsed textual embeddings.
Queries remain textual but are projected into the shared multimodal embedding space, where cross-modal similarity is computed to retrieve the most relevant pages.
This task directly measures a model’s ability to bridge textual queries with visual document representations—a key capability for real-world retrieval scenarios in which crucial information is frequently embedded within complex visual structures.

\subsection{Evaluation Metrics}
The performance of each retrieval model is evaluated using two standard metrics widely adopted in information-retrieval research: Recall and Normalized Discounted Cumulative Gain (nDCG). Recall@k measures whether the relevant document appears within the top-k retrieved results, thereby assessing the model’s coverage of correct answers. It serves as an indicator of whether the retrieval system successfully includes the ground-truth document among its top-ranked candidates. nDCG@k evaluates the ranking quality of the retrieved list by assigning higher scores when the correct document is ranked closer to the top. In cases where each query has a single ground-truth answer, nDCG effectively reflects how early that correct page appears in the ranked list, providing a more fine-grained assessment than Recall alone.
This metric is particularly useful when comparing models that may all retrieve the correct page but differ in their ability to rank it precisely.

\subsection{Baseline Models}
The SDS KoPub VDR benchmark comprehensively compares three categories of retrieval models — representative text embedding models, multimodal embedding models, and a custom model developed in this study — to evaluate performance across diverse retrieval scenarios ranging from text-only to visually grounded document understanding. This comparative analysis provides a foundation for quantitatively examining the relative strengths and limitations of classical text-based retrieval and modern multimodal embedding approaches.

\subsubsection{Text Embedding Models}
For Task 1 (Text-only Retrieval), we selected a diverse set of representative text embedding models based on their multilingual capabilities, efficiency, and prevalence in commercial applications. While Korean-specific models exist, our selection prioritizes state-of-the-art multilingual models that have demonstrated strong performance on Korean benchmarks, reflecting the current trend toward universal text representations.

\begin{itemize}
\item BGE-M3 \cite{chen2024bgem3embeddingmultilingualmultifunctionality} (BAAI/bge-m3): A high-performance multilingual embedding model optimized for both dense and lexical retrieval within a single architecture. Its extended 8K-token context window allows for effective processing of long Korean policy documents.
\item Kanana-Nano-2.1B-Embedding \cite{kananallmteam2025kananacomputeefficientbilinguallanguage} (kakaocorp/kanana-nano-2.1b-embedding):  A 2.1B-parameter model developed by Kakao, emphasizing lightweight deployment and computational efficiency. Despite its compact size, it achieves competitive accuracy with significantly lower latency and resource consumption. 
\item Qwen3-Embedding \cite{zhang2025qwen3embeddingadvancingtext} (Qwen/Qwen3-Embedding-0.6B): A text-only embedding model from the Qwen series, designed to capture sentence-level semantics with strong cross-lingual capability, showing particularly robust performance on non-English languages such as Korean.
\item OpenAI Embedding \cite{openai-text-embedding-3-large} (text-embedding-3-large): A widely adopted commercial model known for its consistently strong performance across diverse benchmarks. It employs Matryoshka Representation Learning (MRL) to allow dynamic adjustment of output dimensionality. For our experiments, we set the dimension to 3,072 to balance accuracy and efficiency.
\end{itemize}

These text-embedding baselines rely solely on textual representations for retrieval, serving as reference points for evaluating the contribution of visual and structural cues in multimodal settings.

\subsubsection{Multimodal Embedding Models}
For Task 2 (Multimodal Retrieval), we evaluated models capable of jointly processing visual and textual information from document pages.

\begin{itemize}
\item DSE-Qwen2-2b-MRL-V1 \cite{ma2024unifyingmultimodalretrievaldocument} (MrLight/dse-qwen2-2b-mrl-v1): A Document Screenshot Embedding (DSE) model specialized in preserving visual document layouts. It encodes raw screenshot images directly—without requiring any OCR—thereby integrating textual, visual, and structural cues into a unified representation.
\item Nomic-Embed-Multimodal-7B \cite{nomicembedmultimodal2025} (nomic-ai/nomic-embed-multimodal-7b): A large-scale model that encodes mixed text-image inputs through a single unified encoder. It eliminates the need for separate preprocessing pipelines and can seamlessly handle richly formatted document pages.
\item Jina-Embeddings-v4 \cite{jinaembeddingsv4universalembeddingsmultimodal} (jinaai/jina-embeddings-v4): A dual-mode model supporting both single-vector (dense) and multi-vector (ColBERT-style) retrieval. For consistency across baselines, our experiments adopt the single-vector configuration, emphasizing scalability and inference speed.
\end{itemize}

Unlike the text-only models, these multimodal encoders exploit page-level visual features—such as layout, charts, and tables—allowing the benchmark to assess retrieval performance when structural and graphical information is available. 

Finally, we evaluated SDS-Multimodal-Embedding-7B, a model developed for this work. This model was created by fine-tuning Qwen2.5-VL-7B on a custom-collected dataset of Korean public documents using a multi-stage fine-tuning strategy. The SDS-Multimodal-Embedding-7B model was tested under both retrieval configurations. All baseline and custom models were evaluated under a unified embedding-generation and retrieval protocol, as detailed in Section 5.3.3.

\subsubsection{Evaluation Protocol}
\noindent\textbf{Embedding Generation.}\hspace{0.5em}
For each retrieval task, models converted queries and candidate pages into fixed-dimensional vector representations according to their respective modality configurations. 
Each model preserved its native tokenizer and preprocessing pipeline from pre-training to maintain semantic alignment. 
All embeddings were generated in inference mode on a single GPU, with batch caching and gradient updates disabled to prevent memory-based bias. 
For multimodal models, the output embedding was taken from the final pooled vector or the Last token, depending on the model architecture.

\vspace{0.5em}
\noindent\textbf{Indexing and Similarity Computation.}\hspace{0.5em}
All page embeddings were stored in a FAISS index to enable efficient similarity search. 
Cosine similarity was adopted as the primary metric for retrieval. 
During evaluation, each query embedding was compared against all indexed document embeddings within FAISS, and the top-10 nearest neighbors were retrieved based on similarity scores.

\vspace{0.5em}
\noindent\textbf{Implementation Details.}\hspace{0.5em}
All experiments were conducted on NVIDIA A100 (80 GB) GPUs using PyTorch 2.6. 
Embedding dimensionalities (typically 1K–4K) were kept identical to each model’s native configuration, without any projection or fine-tuning. To ensure consistency, the batch size was fixed at 12, and normalization and mixed-precision (bfloat16) settings were applied uniformly across all experiments to minimize external variance.

\subsection{Results}
In this section, we compare the performance of the aforementioned models on Task 1 (Text-only Retrieval) and Task 2 (Multimodal Retrieval). We further conduct domain-wise and query-type-wise analyses to discuss the distinct characteristics and limitations of each approach.

\subsubsection{Task 1: Text-only Retrieval Results}
\begin{figure}[htb]
  \centering
  % (a) Table
  \begin{subfigure}{\linewidth}
    \centering
    \label{fig:textonly:table}
    \resizebox{\linewidth}{!}{
      \begin{tabular}{lcccccccc}
        \toprule
        \multicolumn{1}{c}{Model} & \multicolumn{4}{c}{Recall@k} & \multicolumn{4}{c}{nDCG@k} \\
        \cmidrule(lr){2-5} \cmidrule(lr){6-9}
        & @1 & @3 & @5 & @10 & @1 & @3 & @5 & @10 \\
        \midrule
        BGE-M3                            & 0.41 & 0.68 & 0.75 & 0.82 & 0.41 & 0.46 & 0.49 & 0.57 \\
        Kanana-Nano-2.1B-Embedding        & 0.46 & 0.66 & 0.74 & 0.81 & 0.46 & 0.50 & 0.53 & 0.59 \\
        Qwen3-Embedding-0.6B              & 0.38 & 0.60 & 0.68 & 0.78 & 0.38 & 0.43 & 0.47 & 0.54 \\
        text-embedding-3-large (OpenAI)   & 0.40 & 0.64 & 0.72 & 0.81 & 0.40 & 0.45 & 0.49 & 0.56 \\
        Jina-Embeddings-v4                & 0.49 & 0.71 & 0.79 & 0.85 & 0.49 & 0.54 & 0.57 & 0.64 \\
        SDS-Multimodal-Embedding-7B       & 0.54 & 0.77 & 0.83 & 0.89 & 0.54 & 0.58 & 0.62 & 0.68 \\
        \bottomrule
      \end{tabular}
    }
  \end{subfigure}

  \vspace{0.8em}

  % (b) Plot subfigure
  \begin{subfigure}{1.0\linewidth}
    \centering
    \includegraphics[width=\linewidth]{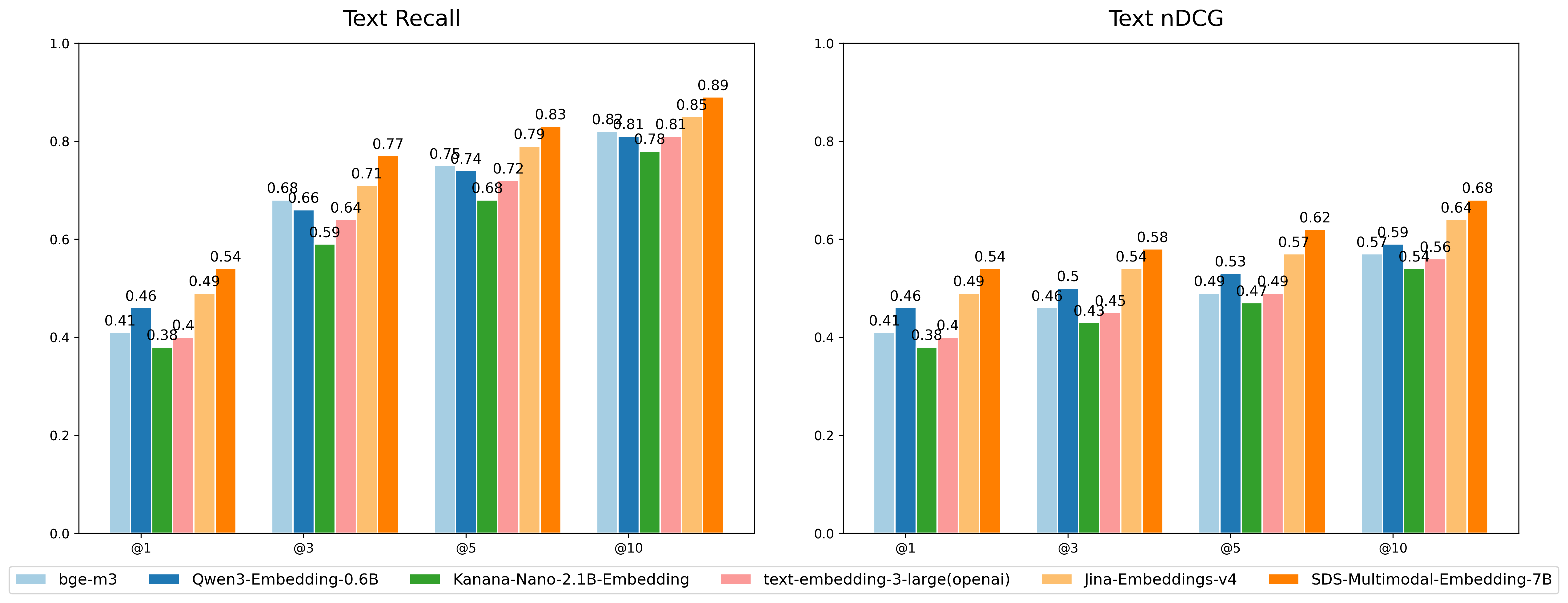}
    \label{fig:textonly:plot}
  \end{subfigure}

  \caption{Summary of text-only retrieval performances}
  \label{fig:textonly}
\end{figure}

In Task 1, we evaluated retrieval performance based on the textual content of documents, measuring similarity between query and passage embeddings. The results are summarized in Figure \ref{fig:textonly}.
Among the evaluated models, only SDS-Multimodal-Embedding-7B and Jina-Embeddings-v4 were trained with multimodal objectives; all others were text-only embedding models. 

Among the text-only baselines, BGE-M3 exhibited the most consistent and stable performance, showing particularly strong results on explicit keyword-based queries. The Kanana-Nano-2.1B-Embedding, despite its compact size, maintained relatively high accuracy, demonstrating a well-balanced trade-off between efficiency and retrieval precision. Interestingly, Jina-Embeddings-v4, although originally a multimodal model, performed competitively even when evaluated using text-only indexing, achieving Recall@3 = 0.71. This suggests that its large-scale vision-language pretraining contributes to enhanced semantic diversity and robustness in its textual representations, even when the visual modality is not directly used. 

In contrast, our SDS-Multimodal-Embedding-7B, fine-tuned specifically on public documents, achieved robust retrieval performance even for complex contextual queries. It recorded Recall@3 = 0.77, representing a 13.24\% improvement over BGE-M3. This result demonstrates that domain-adapted multimodal pretraining, when aligned with structured and context-rich government documents, substantially improves the model’s ability to retrieve semantically relevant content beyond surface-level keyword matching.

\subsubsection{Task 2: Multimodal Retrieval Results}

\begin{figure}[htb]
  \centering

  % (a) Table
  \begin{subfigure}{\linewidth}
    \centering
    \label{fig:combo:table}
    \resizebox{\linewidth}{!}{
      \begin{tabular}{lcccccccc}
        \toprule
        \multicolumn{1}{c}{Model} & \multicolumn{4}{c}{Recall@k} & \multicolumn{4}{c}{nDCG@k} \\
        \cmidrule(lr){2-5} \cmidrule(lr){6-9}
        & @1 & @3 & @5 & @10 & @1 & @3 & @5 & @10 \\
        \midrule
        dse-qwen2-2b-mrl-v1         & 0.23 & 0.40 & 0.46 & 0.54 & 0.23 & 0.27 & 0.29 & 0.35 \\
        Nomic-Embed-Multimodal-7B   & 0.47 & 0.67 & 0.74 & 0.83 & 0.48 & 0.52 & 0.55 & 0.61 \\
        Jina-Embeddings-v4          & 0.46 & 0.66 & 0.74 & 0.82 & 0.46 & 0.50 & 0.54 & 0.60 \\
        SDS-Multimodal-Embedding-7B & 0.63 & 0.86 & 0.90 & 0.95 & 0.63 & 0.67 & 0.70 & 0.76 \\
        \bottomrule
      \end{tabular}
    }
  \end{subfigure}

  \vspace{0.8em}

  % (b) Plot subfigure
  \begin{subfigure}{1.0\linewidth}
    \centering
    \includegraphics[width=\linewidth]{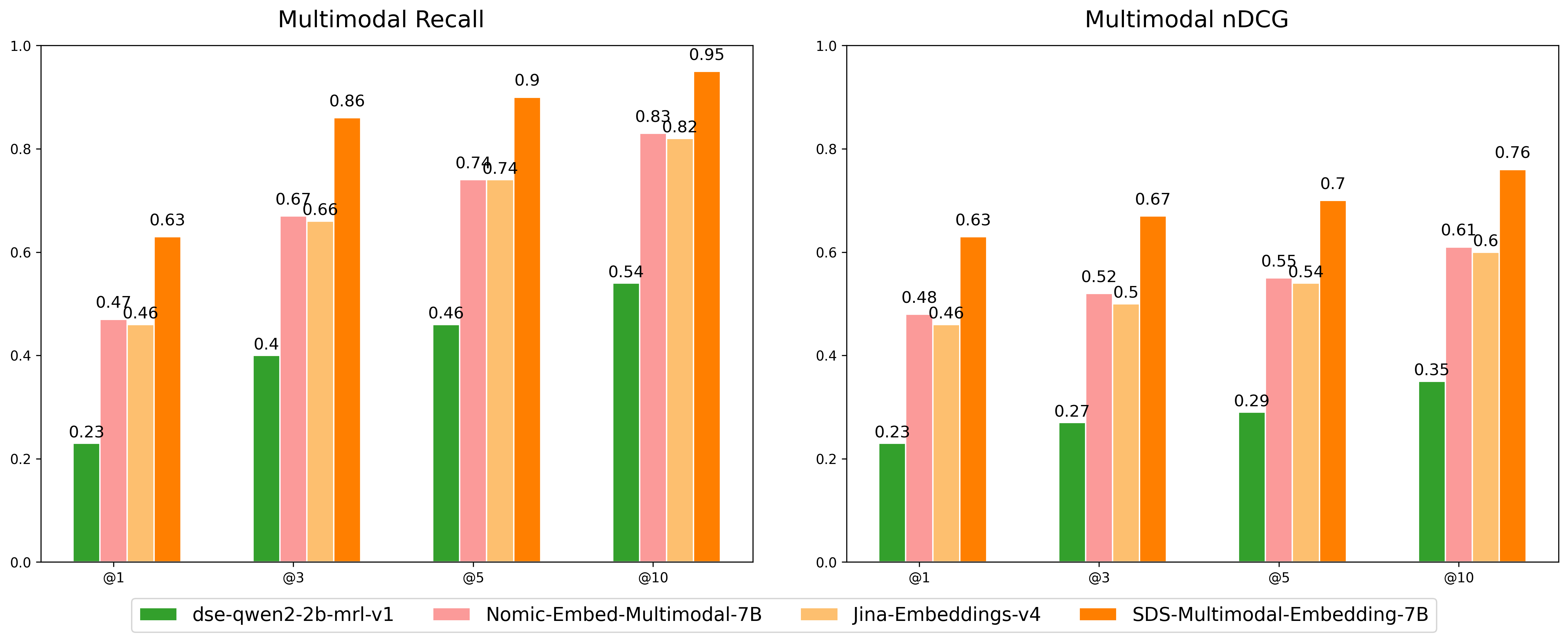}
    \label{fig:textonly:plot}
  \end{subfigure}

  \caption{Summary of multimodal retrieval performances}
  \label{fig:multimodal_results}
\end{figure}

In multimodal retrieval, we evaluate performance based on the similarity between the document image and the query text. The experimental results are shown in Figure \ref{fig:multimodal_results}. The Nomic-Embed-Multimodal and Jina-Embeddings-v4 models (based on Qwen2.5-VL-7B-Instruct and Qwen2.5-VL-3B-Instruct, respectively) exhibit minor differences based on model size but consistently outperform all text-only models. This suggests that even for models lacking specific pre-training in Korean, the image modality can provide a more accurate representation of a document's content. Our model, fine-tuned from a Qwen2.5-VL-7B base on our public document dataset, surpasses the Nomic-Embed-Multimodal-7B by a significant margin of over 21\% in Recall@5. Critically, when comparing Task 1 and Task 2 under an identical architecture (i.e., changing only the input modality), multimodal-based retrieval improves upon text-based retrieval by 8.4\% in Recall@5. Qualitative analysis reveals that this performance gain is primarily attributed to the model's ability to accurately recognize and interpret visual elements, such as numerical values in tables, graph legends, and image captions. These findings underscore the necessity of a multimodal approach, particularly for visually complex public documents.

\subsubsection{Domain-wise Performance Analysis}

\begin{figure}[ht]
  \centering
  % (a) Left image
  \begin{subfigure}{0.6\textwidth}
    \centering
    \includegraphics[width=\linewidth]{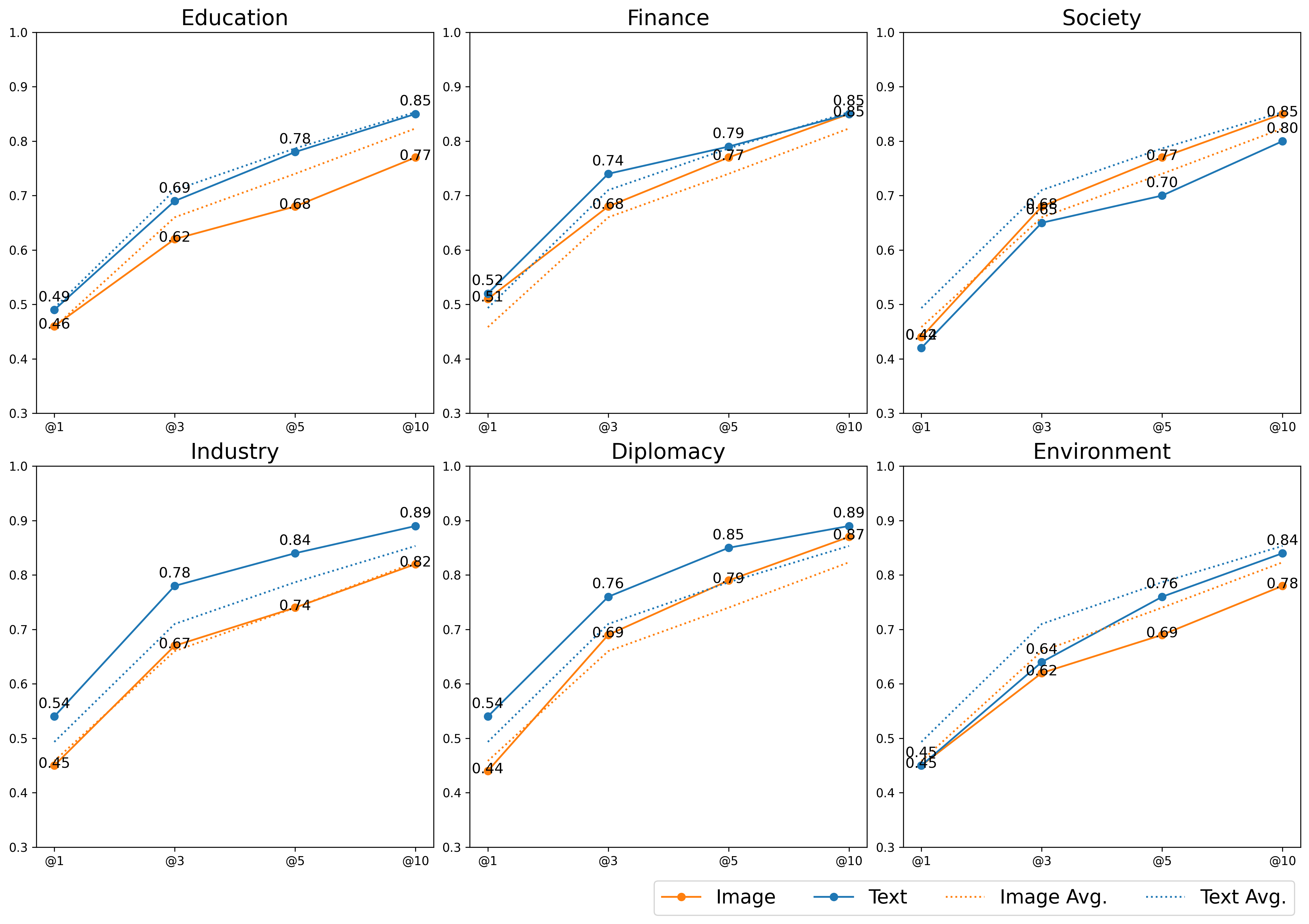}
    \caption{Jina-Embeddings v4}
    \label{fig:domain_aware_text}
  \end{subfigure}
  \hfill
  % (b) Right image
  \begin{subfigure}{0.6\textwidth}
    \centering
    \includegraphics[width=\linewidth]{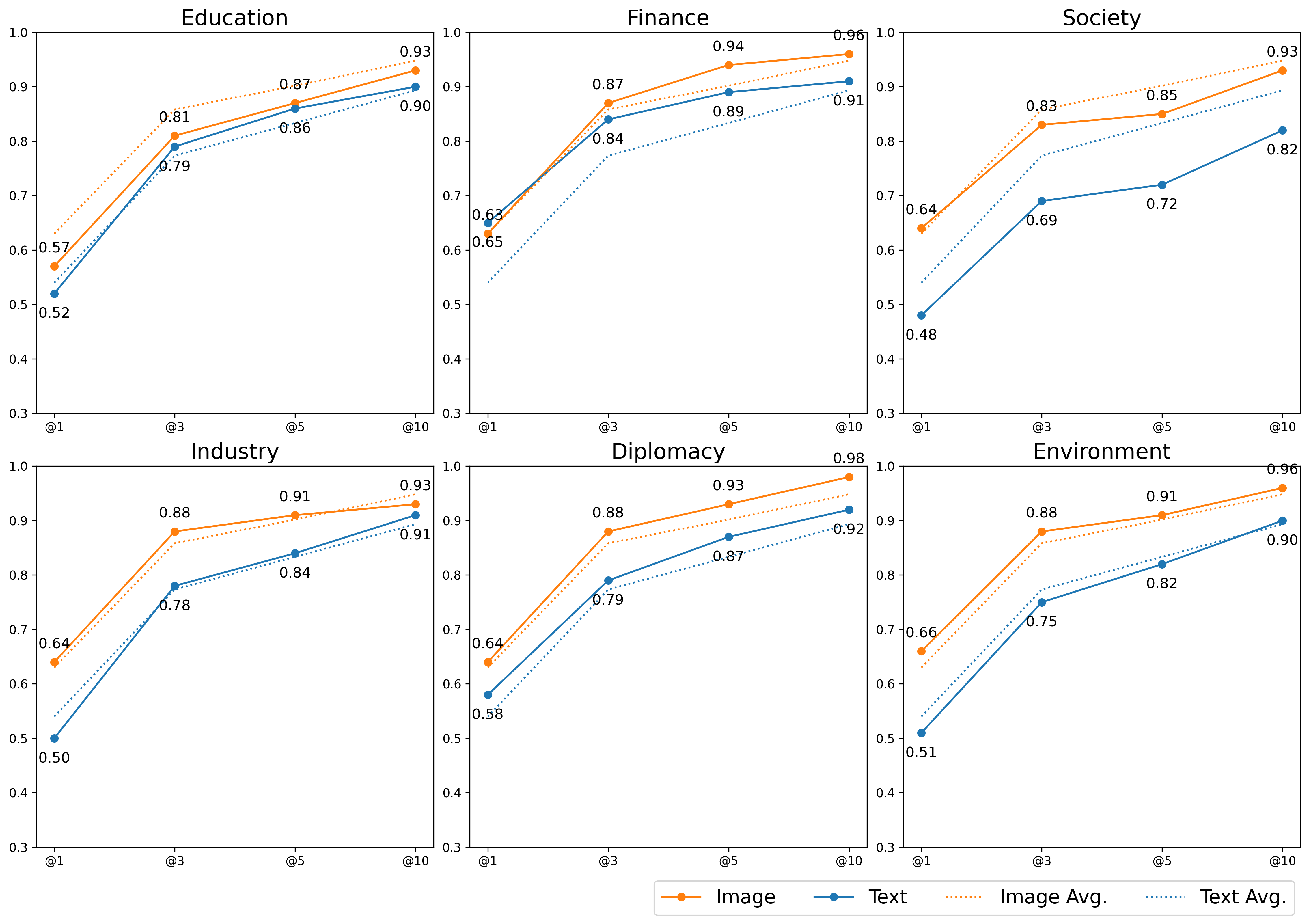}
    \caption{SDS-Multimodal-Embedding-7B}
    \label{fig:domain_aware_image}
  \end{subfigure}

  \caption{Domain-wise retrieval performance (Recall).}
  \label{fig:domain_aware}
\end{figure}

We conducted a domain-wise analysis to compare performance variations across modalities. The evaluation involved two representative models:
(i) SDS-Multimodal-Embedding-7B, a domain-adapted model fine-tuned on Korean public administrative documents, and (ii) Jina-Embeddings-v4, a general-purpose model trained on large-scale multimodal representations without domain-specific supervision.

For each model, we independently constructed text-based and image-based embeddings from identical document pages and measured domain-wise Recall@k scores. The results are illustrated in Figure \ref{fig:domain_aware}.

Overall, SDS-Multimodal-Embedding-7B consistently achieved the highest performance across all domains. In image-based retrieval, SDS-Multimodal-Embedding-7B exhibited strong performance starting from Recall@1 between 0.57 and 0.66 across major domains such as Education (0.57), Finance (0.63), and Environment (0.66), with Recall@10 increasing steadily to the 0.93 – 0.98 range.
In domains that heavily rely on structured visual data — such as Finance, Diplomacy, and Environment — the model maintained particularly high accuracy (e.g., Recall@10 = 0.98 for Diplomacy). These results indicate that SDS-Multimodal-Embedding-7B effectively captures visual contextual cues such as color-coded legends, axis labels, and directional arrows — information often unavailable from pdf-extracted text alone.

Jina-Embeddings-v4 also demonstrated stable and coherent domain-wise performance, with relatively stronger results in Finance and Diplomacy.
For instance, its image-based retrieval achieved Recall@1 = 0.51 and Recall@10 = 0.85 in Finance, and Recall@1 = 0.44 to Recall@10 = 0.87 in Diplomacy. Although its retrieval scores were lower than SDS-Multimodal-Embedding-7B, Jina-Embeddings-v4’s consistency across heterogeneous domains is noteworthy, given that it relies solely on general pretraining without any domain adaptation. This finding suggests that large-scale multimodal pretraining can transfer reasonably well even to public administrative documents, providing baseline retrieval capability in the absence of specialized finetuning.

From a modality perspective, both models exhibited domain-dependent preferences —that is, the dominant modality varied by domain. For SDS-Multimodal-Embedding-7B, image-based retrieval clearly outperformed text-based retrieval in visually structured domains such as Finance, Diplomacy, and Environment, where tables, charts, and time-series plots constitute the main carriers of semantic meaning. In the Diplomacy and Environment domains, image-based Recall curves were comparable to or even exceeded their text-based counterparts, reflecting that visual representations often serve as the primary evidential source (e.g., quantitative tables, labeled diagrams). This pattern was especially pronounced in the Environment domain, which features a dense mix of diagrams, icons, and color-coded visual markers. Conversely, Education exhibited smaller modality gaps, as its documents predominantly consist of explanatory paragraphs and repetitive terminologies, indicating that text alone is often sufficient in linguistically homogeneous domains.

A similar trend was observed for Jina-Embeddings-v4. In visually structured domains (Finance, Diplomacy, Industry), image-based retrieval achieved Recall@1 scores of 0.44 – 0.51. In contrast, for text-centric domains (Social and Education), text-based retrieval started lower at Recall@1 but improved steadily with higher k, suggesting that Jina’s text representations capture the broader semantic context even without explicit domain tuning. Nonetheless, when queries required identifying visually grounded elements (e.g., metric names within tables, legend values, trend markers), the multimodal embeddings offered a clear advantage, implying that the model’s vision-language alignment provides transferable grounding even in unseen document types.

In summary: (1) At the domain level, multimodal embeddings proved particularly effective in domains where visual structures convey core information — notably Finance, Diplomacy, and Environment. (2) At the model level, the domain-adapted SDS-Multimodal-Embedding-7B achieved higher accuracy and stronger cross-modal alignment compared to the general-purpose Jina-Embeddings-v4. These results collectively underscore that text-only embeddings are insufficient for public administrative and policy documents, where tabular and graphical elements often carry essential semantics. Moreover, the results from Jina-Embeddings-v4 reveal that even without fine-tuning, general multimodal pretraining provides a solid foundation, suggesting that domain-specific post-adaptation of such general models could potentially approach the specialized performance of SDS-Multimodal-Embedding-7B.

\subsubsection{Query Type-wise Performance Analysis}

\begin{figure}[ht]
  \centering

  % (a) Top subfigure
  \begin{subfigure}{\linewidth}
    \centering
    \includegraphics[width=0.9\textwidth]{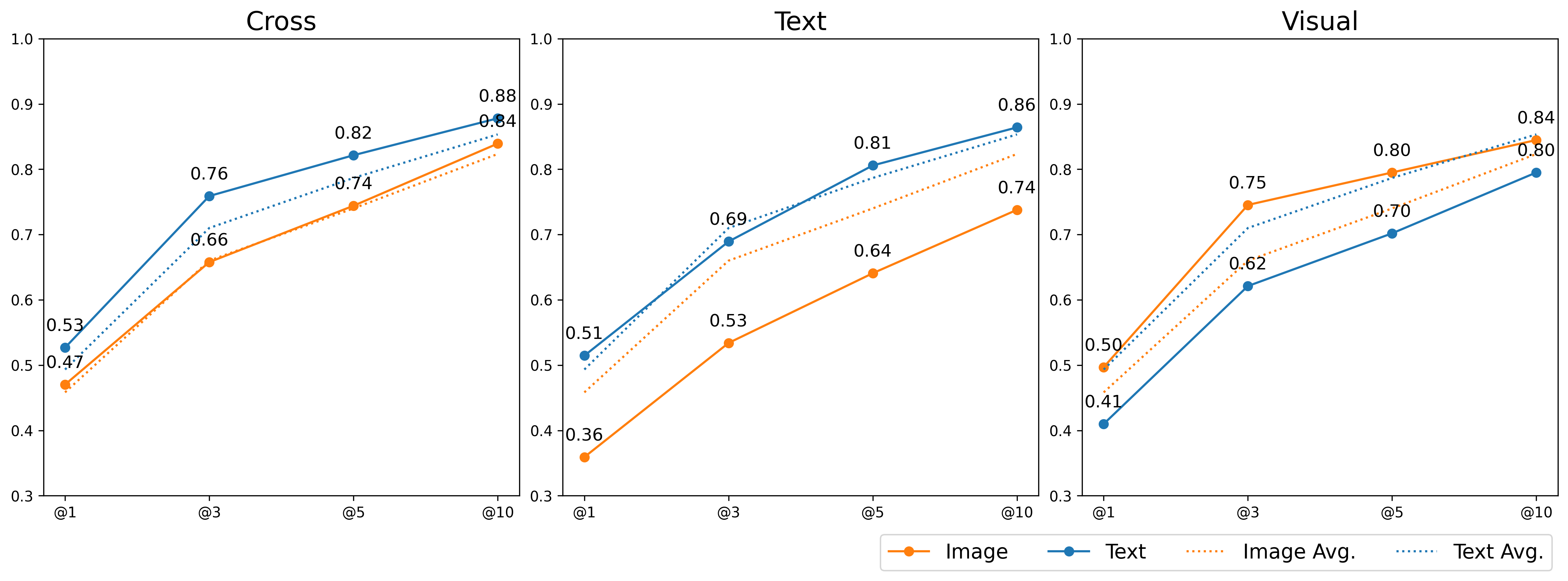}
    \caption{Jina Embedidngs v4}
    \label{fig:query_type_aware_text}
  \end{subfigure}

  \vspace{0.8em}

  % (b) Bottom subfigure
  \begin{subfigure}{\linewidth}
    \centering
    \includegraphics[width=0.9\textwidth]{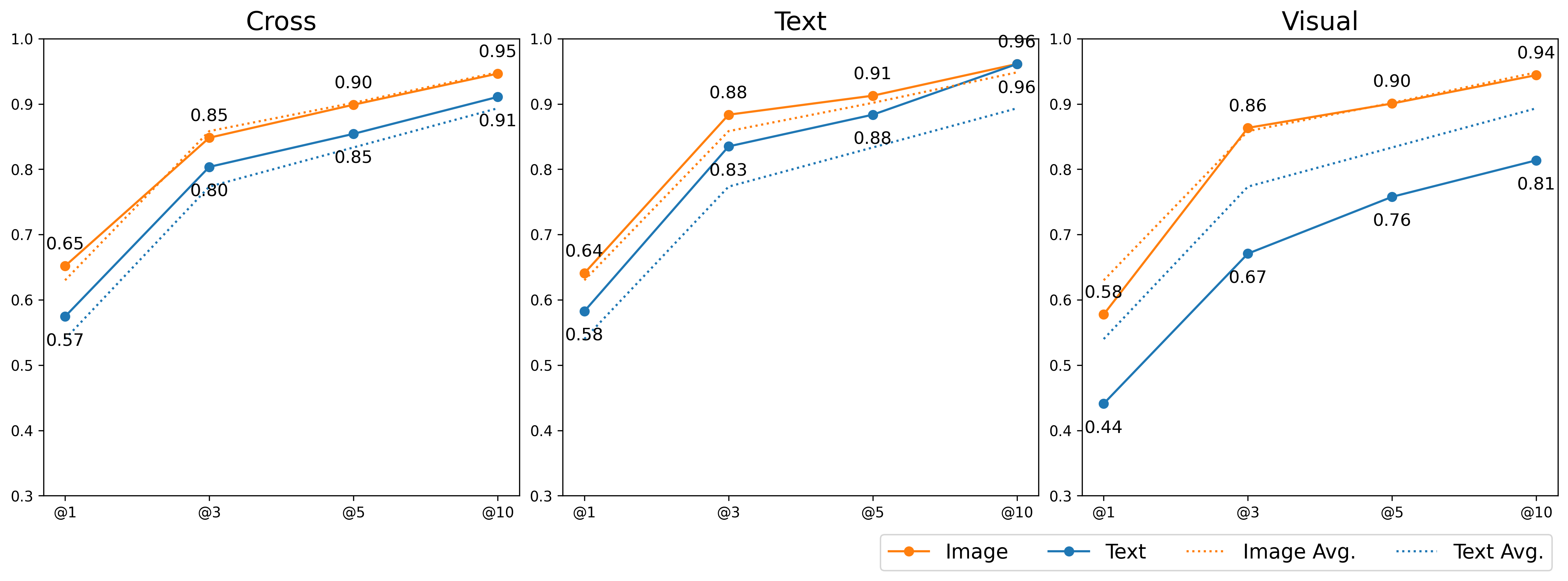}
    \caption{SDS-Multimodal-Embedding-7B}
    \label{fig:query_type_aware_image}
  \end{subfigure}

  \caption{Query-type specific retrieval performance (Recall)}
  \label{fig:query_type_aware}
\end{figure}

To analyze retrieval performance based on query type, we used the SDS-Multimodal-Embedding-7B and Jina-Embeddings v4 models, consistent with Section 5.4.3. We compared Recall@k for Textual, Visual, and Cross queries (refer to Figure \ref{fig:query_type_aware}).

The most prominent difference emerged in Visual queries. These queries directly depend on visual elements within a document—such as tables, graphs, and diagrams—and often cannot be adequately resolved using textual information alone. For this query type, the multimodal index served as the key driver of performance enhancement for both models.

First, for SDS-Multimodal-Embedding-7B, the application of a multimodal index achieved a Recall@1 of 0.58, Recall@3 of 0.86, and Recall@10 of 0.94 on Visual queries. This represents a significant improvement over the model's text-only index, with gains of approximately +28\%p at Recall@3 (0.86 vs. 0.67) and +16\%p at Recall@10 (0.94 vs. 0.81). This enhancement stems from the model's ability to directly recognize and utilize evidence present only in visual elements, such as legend colors, axis units, or arrows indicating trends. This demonstrates that, for queries where the visual context constitutes the answer itself, a multimodal index is effectively indispensable. Given that graphs and tables in public documents often carry more critical information than the surrounding text, the direct referencing of this visual data substantially strengthens the model's practical reasoning capabilities. Jina-Embeddings-v4 exhibited a similar trend on Visual queries. While its text-only index (Jina-text) stalled at a Recall@1 of 0.41 and Recall@10 of 0.80, its multimodal index (Jina-multimodal) improved these scores to 0.50 (Recall@1) and 0.84 (Recall@10). This equates to a performance gain of approximately +14\%p to +22\%p when applying multimodal indexing. This result implies that Jina-Embeddings-v4 possesses an inherent capability to interpret visual cues derived solely from its general-purpose pre-training. Although it lacks domain-specific adaptation, it has secured a sufficiently generalizable expressive power for visual information processing. Notably, the Jina-Embeddings-v4 model tends to reinforce visual meaning by leveraging pdf-extracted text (such as legends and titles) from structured images, suggesting a successful transfer of its foundational visual-language alignment.

In summary, both SDS-Multimodal-Embedding-7B and Jina-Embeddings-v4 demonstrated that a multimodal index holds a distinct advantage over a text-only index for Visual queries. This was a common phenomenon observed irrespective of model architecture or training data. It suggests that when visual representations within a document serve as the core informational cues, the indexing structure that enables access to visual information becomes a determining factor in retrieval accuracy. SDS-Multimodal-Embedding-7B, specialized for the visual structures of public documents, exhibited a larger margin of improvement. Jina-Embeddings-v4, despite its general-purpose foundation, demonstrated a competent baseline for interpreting visual cues. These results validate that Visual queries are the most effective type for verifying the utility of multimodal models. Consequently, in environments involving public documents rich with visual information, the limitations of text-based retrieval are evident, confirming that the adoption of a multimodal index is a prerequisite for achieving substantial performance gains.

\subsubsection{Comparative Discussion}
The preceding results demonstrate that the SDS KoPub VDR benchmark is a robust tool for the multidimensional evaluation of retrieval models. While text-based approaches remain useful for explicit queries, a multimodal approach is indispensable in document environments with complex visual structures. In particular, our model, developed based on Qwen2.5-VL-7B, shows a clear advantage in semantic alignment quality and structural understanding over existing models, significantly advancing the performance of multimodal retrievers. These findings open up diverse future research directions, including multimodal representation learning, optimization of vision-language alignment, and the design of next-generation multimodal RAG systems.

\section{Discussion and Conclusion}
In this study, we introduce SDS KoPub VDR, the first VDR benchmark in Korea designed to systematically evaluate an AI's ability to deeply understand and retrieve complex visual information embedded in public documents. Traditional text-centric evaluation methods have clear limitations, often failing to capture critical information encoded in tables, charts, and complex layouts. SDS KoPub VDR addresses this gap by leveraging real-world public documents and incorporating textual, visual, and cross-modal queries. It is significant in providing a standardized evaluation framework to assess the performance of next-generation technologies like Multimodal RAG and domain-specific Vertical AI systems. 

While this benchmark lays a critical foundation for VDR research, it also has several limitations that clearly define directions for future work. 

First is the dataset's scale and domain diversity. The current dataset, comprising 600 question-answer pairs across six domains, is effective for validating baseline model performance but is insufficient for fine-grained analysis of model robustness or for evaluating generalization across a wide array of public sectors. Furthermore, while the use of LLMs for query generation ensures consistency, it carries a potential limitation: the generated queries may not fully capture the diversity and unpredictable nature of colloquial queries made by real users. To overcome these limitations, future work will focus on significantly expanding the dataset to thousands of pairs and incorporating new key domains such as healthcare, technology, culture, and defense. We will also explore methods to enhance query realism and diversity, such as using crowdsourcing or actual user logs. 

Second is the complexity of the tasks. The current benchmark primarily focuses on single-hop queries, where the answer is located within a single page. However, real-world scenarios in public administration and policy analysis frequently require synthesizing and reasoning over information scattered across multiple pages or even different documents. This challenge is compounded by real-world noise, such as low-quality OCR and non-standard document layouts. Therefore, the benchmark will be advanced to include more complex multi-hop tasks, such as cross-page reasoning and multi-document QA. This will serve as a more rigorous testbed for evaluating a model's higher-order capabilities for information synthesis and complex reasoning. 

In addition to these expansion plans, a long-term goal of this research is to foster a healthy research ecosystem. To this end, we plan to establish a public leaderboard with a standardized evaluation pipeline to encourage community participation and support transparent, reproducible comparisons of model performance. Ultimately, our vision is to evolve the benchmark from its current focus on retrieval to an End-to-End Multimodal RAG evaluation framework that holistically assesses the entire pipeline from retrieving visual evidence to generating accurate answers from it. 

In conclusion, SDS KoPub VDR is not merely a static dataset but a living foundational resource that introduces a new evaluation paradigm for AI research on Korean public documents. By systematically addressing the limitations discussed herein and expanding its scope, we aim to catalyze the development of more robust and reliable AI systems capable of maximizing the value of public data.

\bibliographystyle{unsrtnat}  % 출현 순 + natbib 전용
\bibliography{reference}

\appendix
\section{Appendix}

\subsection{Annotation Schema}
Based on page-level data, we designed an annotation schema structured along three key dimensions: domain classification, query type definition, and answer structure design.
Rather than merely generating QA pairs, this schema enables a quantitative evaluation of how effectively multimodal retrieval models can understand and reason over diverse textual and visual tasks. The schema consists of three hierarchical metadata layers—document-level, page-level, and QA-level—as summarized in Tables \ref{tbl:document-meta}–\ref{tbl:qa-meta}.

\subsubsection{Document-Level Metadata}
At the top level, each document is represented by a unique identifier and its associated metadata, as shown in Table \ref{tbl:document-meta}.
This metadata links each document to its corresponding pages and QA pairs, forming the structural foundation of the benchmark.

\begin{table}[ht]
\centering
\caption{Schema of Document-Level Metadata}
\label{tbl:document-meta}
\begin{tabular}{p{4.5cm} p{2cm} p{8cm}}
\toprule
\textbf{Name} & \textbf{Type} & \textbf{Description} \\
\midrule
\texttt{file\_id} & string & Document ID \\
\texttt{file\_name} & string & Document name \\
\texttt{down\_url} & string & Document's download file link \\
\texttt{page\_indices} & list[] & List of page indices in SDS-KoPub-VDR \\
\texttt{query\_indices} & list[] & List of query–answer set indices in SDS-KoPub-QA \\
\texttt{indication\_of\_the\_source} & string & The source and license of the work \\
\bottomrule
\end{tabular}
\end{table}

\subsubsection{Page-Level Metadata}
Each PDF page serves as the fundamental input unit for multimodal retrieval models. As shown in Table \ref{tbl:page-meta}, the page metadata includes both visual and textual representations extracted during preprocessing, enabling unified multimodal access for retrieval and QA tasks.

\begin{table}[ht]
\centering
\caption{Schema of Page-Level Metadata}
\label{tbl:page-meta}
\begin{tabular}{p{3cm} p{3cm} p{8cm}}
\toprule
\textbf{Name} & \textbf{Type} & \textbf{Description} \\
\midrule
\texttt{id} & string & Identifier for a page ID \\
\texttt{file\_name} & string & Document name \\
\texttt{image} & \texttt{PIL.Image.Image} & PIL image object representing the page image \\
\texttt{text} & string & Text content of the page via PyPDF\\
\texttt{ocr} & string & Text extracted via OCR process \\
\bottomrule
\end{tabular}
\end{table}

\subsubsection{QA-Level Metadata}
Each QA pair forms the basic evaluation unit for both retrieval and QA performance.
As presented in Table \ref{tbl:qa-meta}, the QA-level metadata defines query modality, domain category, and ground-truth evidence for calculating retrieval metrics such as Recall and nDCG.
The \texttt{ground\_truth} field, in particular, plays a key role in verifying whether a model successfully retrieves the correct evidence page.

\begin{table}[htbp]
\centering
\caption{Schema of QA-Level Metadata}
\label{tbl:qa-meta}
\begin{tabular}{p{3cm} p{3cm} p{8cm}}
\toprule
\textbf{Name} & \textbf{Type} & \textbf{Description} \\
\midrule
\texttt{id} & string & Page ID for ground-truth evidence (not unique) \\
\texttt{query} & string & Question text \\
\texttt{answer} & string & Corresponding answer text \\
\texttt{type} & string & The modality type of the question (e.g., Text, Visual, Cross) \\
\texttt{domain} & string & Document's domain or category \\
\texttt{ground\_truth} & list[] & Page indices for ground-truth evidence \\
\bottomrule
\end{tabular}
\end{table}

\subsection{Data Preprocessing Pipeline Example}

\begin{figure}[htbp]
    \centering
    \includegraphics[width=0.9\textwidth]{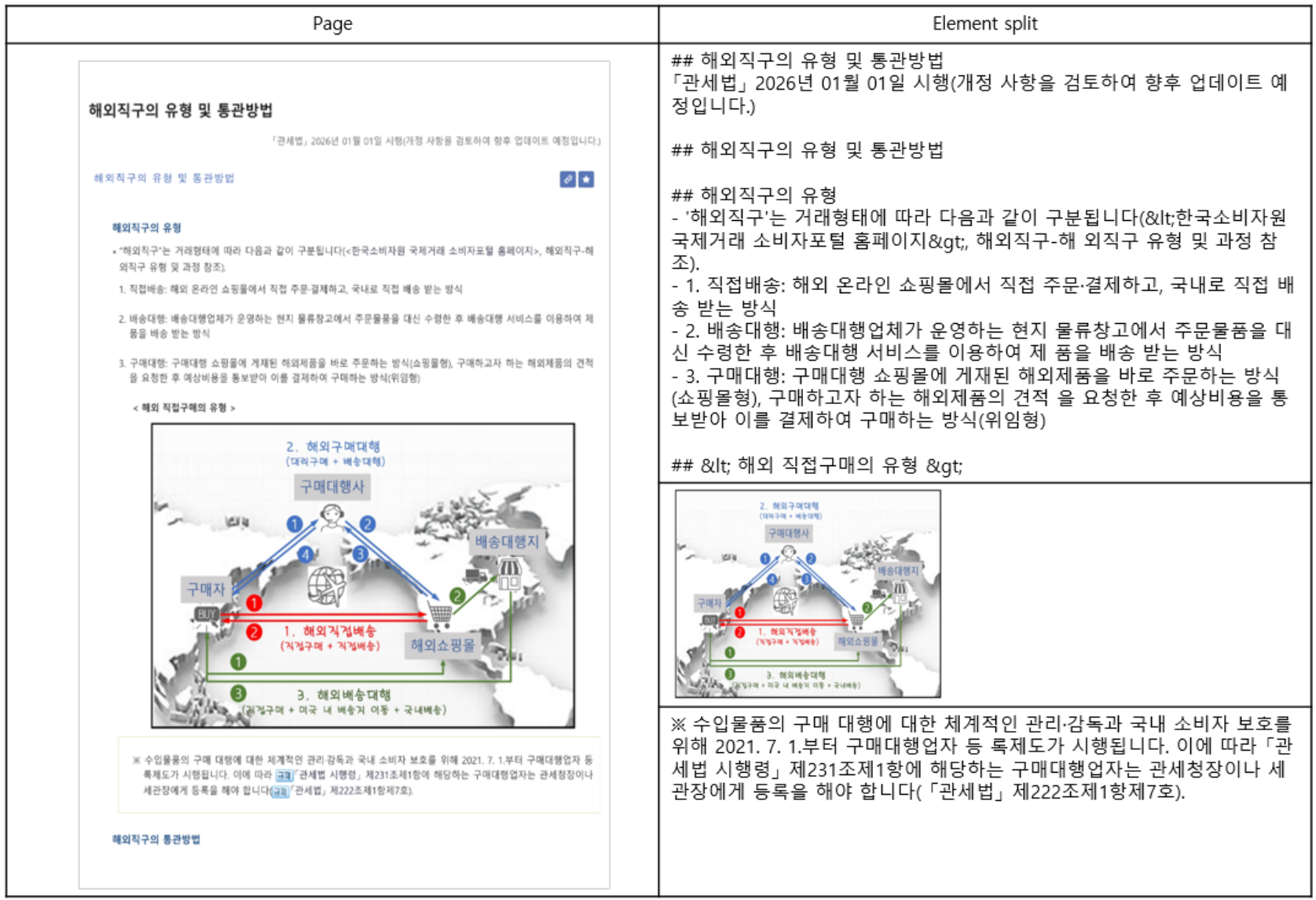}
    \caption{Visual–textual element separation process. 
    The left panel shows the original PDF page, while the right panel presents its decomposed components (text regions and visual elements). 
    The extracted text is used solely as metadata, and visual regions are stored as cropped image patches for later use in multimodal query generation.}
    \label{fig:split_elements}
\end{figure}

\begin{figure}[htbp]
    \centering
    \includegraphics[width=0.9\textwidth]{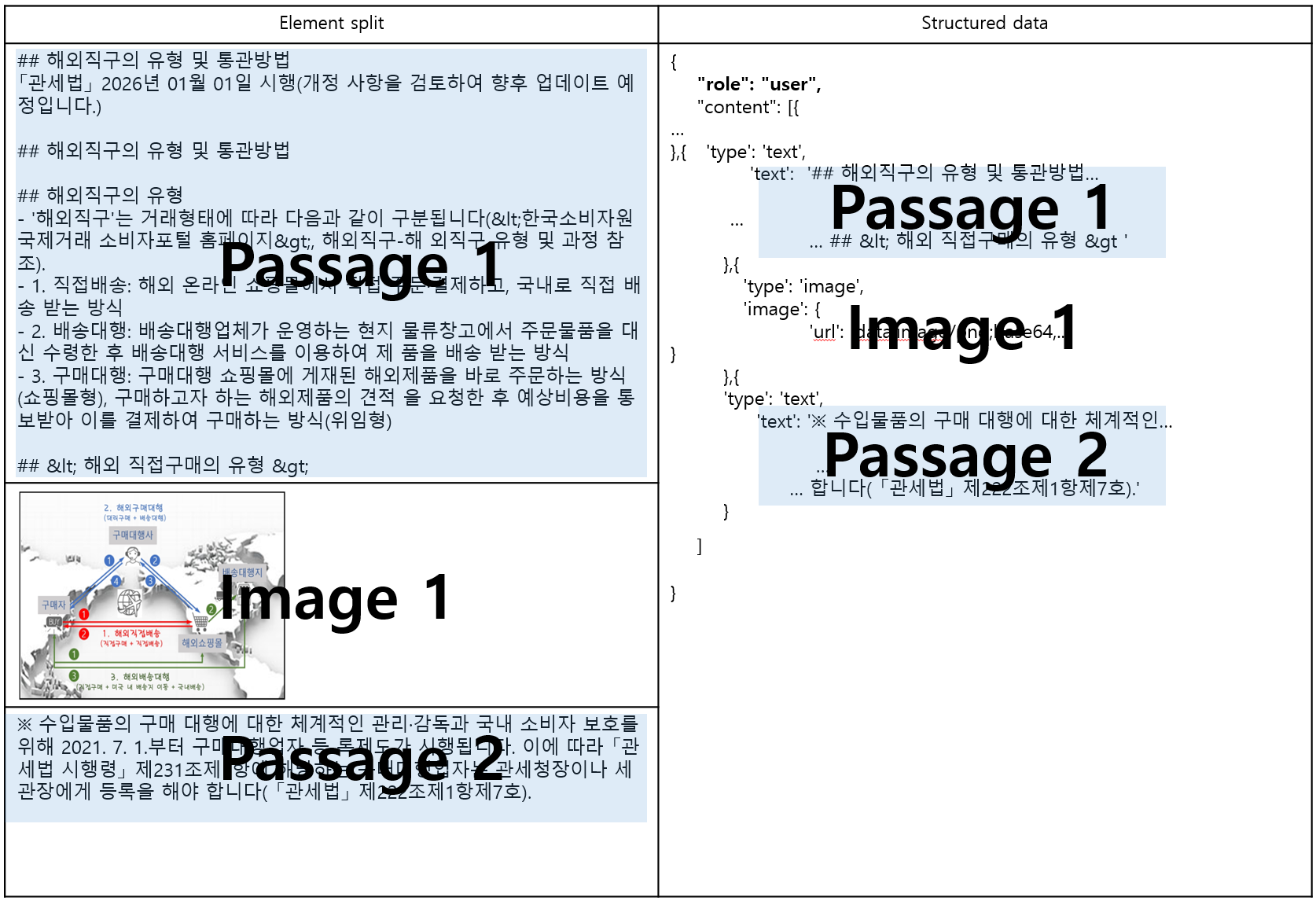}
    \caption{Example of multimodal dataset construction. 
    The left panel corresponds to the separated content from Figure \ref{fig:split_elements}, while the right panel illustrates the structured JSON format used for training. 
    Each entry contains user prompts with distinct \texttt{text} and \texttt{image} fields, enabling fine-grained multimodal alignment.}
    \label{fig:structured_data}
\end{figure}

Figures \ref{fig:split_elements} and \ref{fig:structured_data} visually illustrate the preprocessing and data structuring procedures described in Section 3.2.

\subsection{QA generation Prompts}
\begin{figure}[htbp]
    \centering
    \includegraphics[width=1\textwidth]{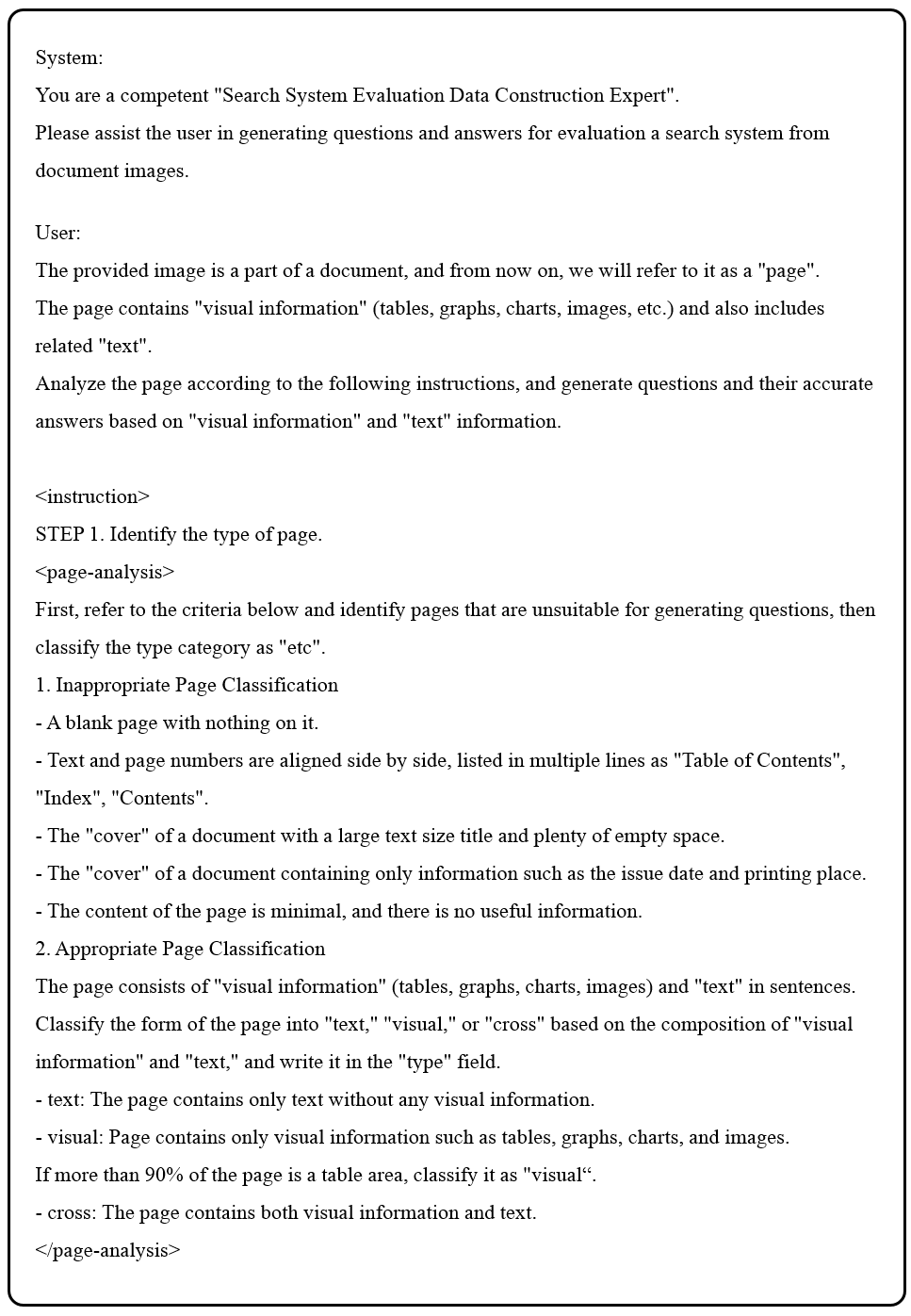}
    \caption{QA generation prompt 1}
    \label{fig:qa_gen_1}
\end{figure}

\begin{figure}[htbp]
    \centering
    \includegraphics[width=1\textwidth]{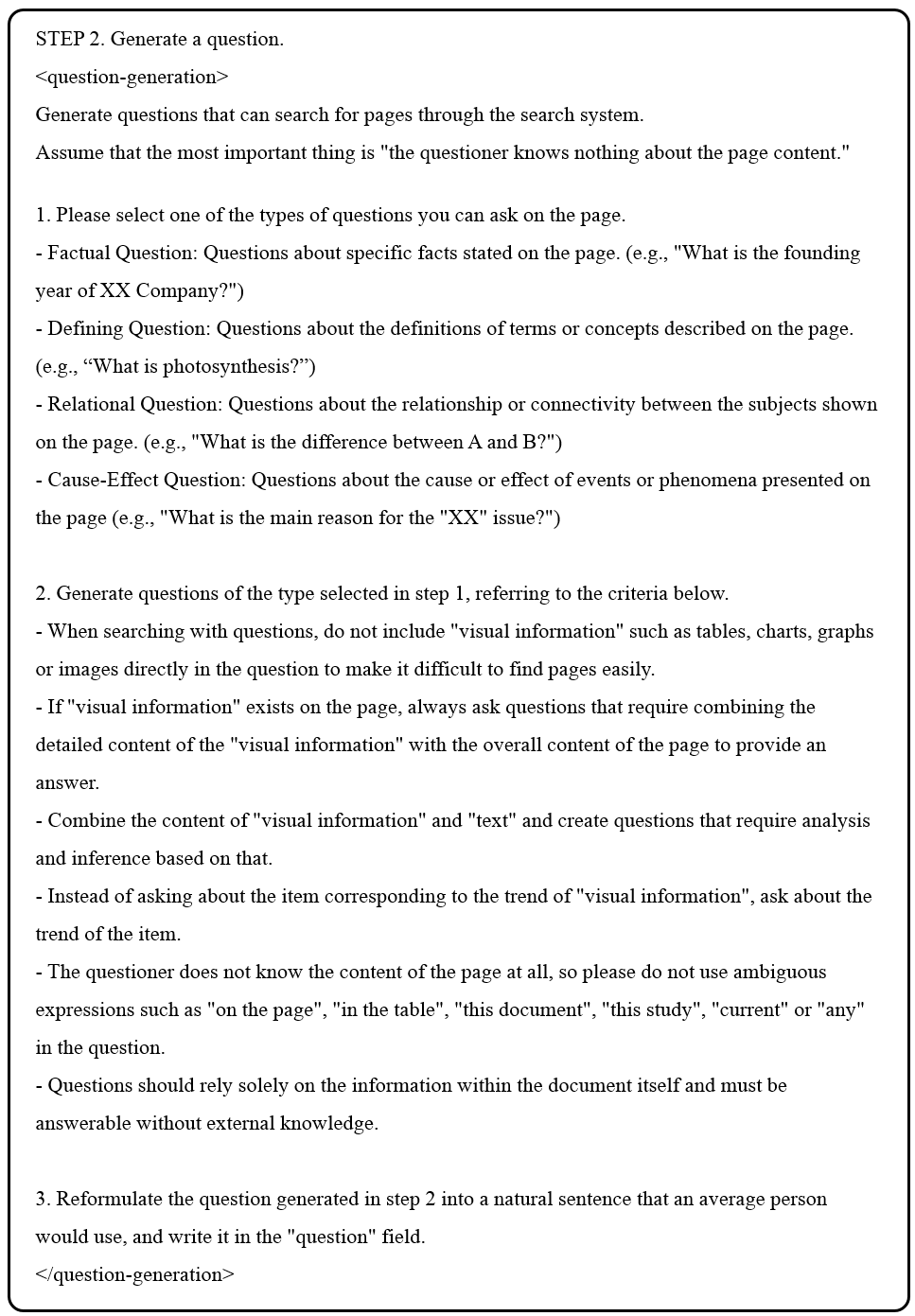}
    \caption{QA generation prompt 2}
    \label{fig:qa_gen_2}
\end{figure}

\begin{figure}[htbp]
    \centering
    \includegraphics[width=1\textwidth]{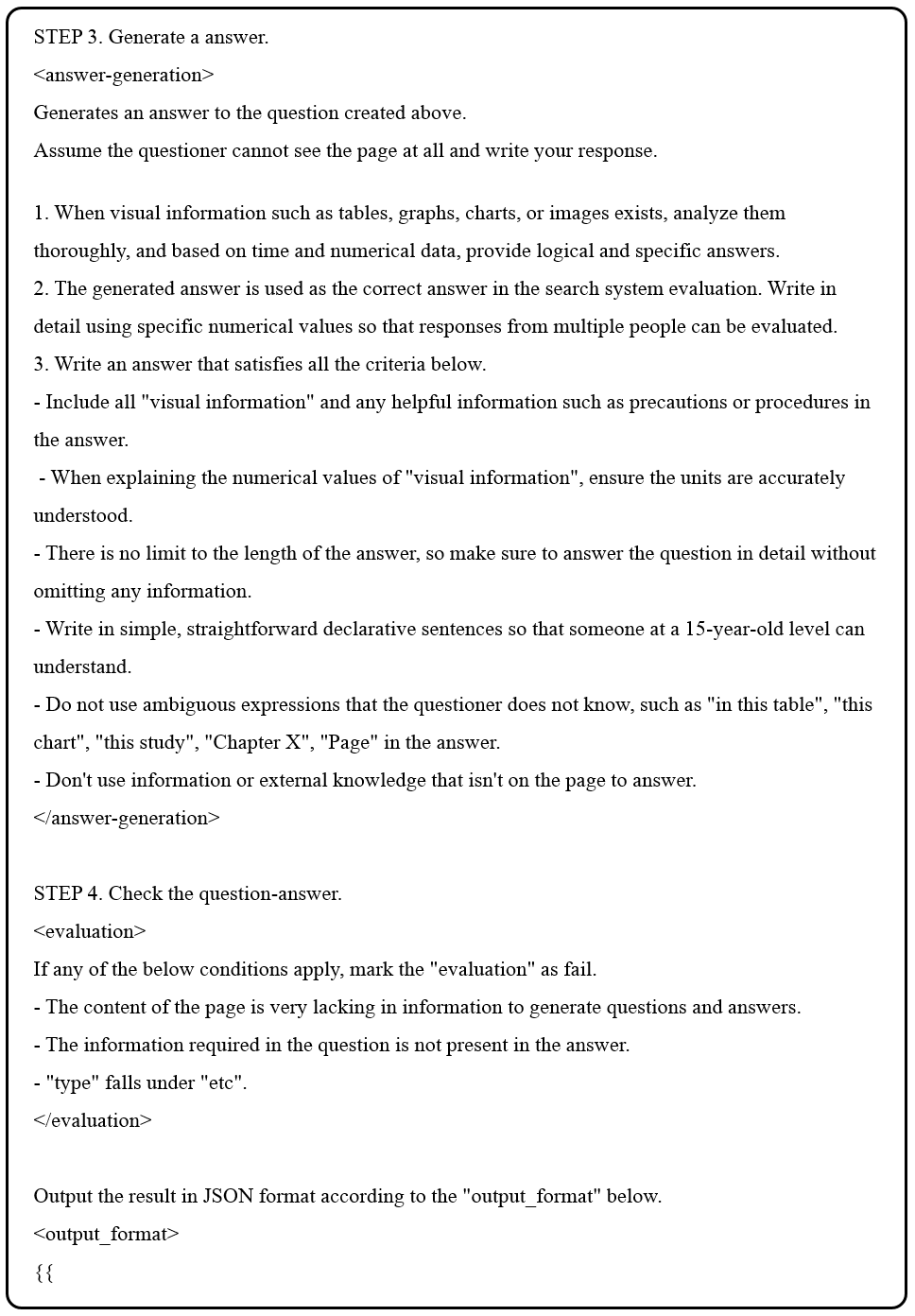}
    \caption{QA generation prompt 3}
    \label{fig:qa_gen_3}
\end{figure}

\begin{figure}[ht]
    \centering
    \includegraphics[width=1\textwidth]{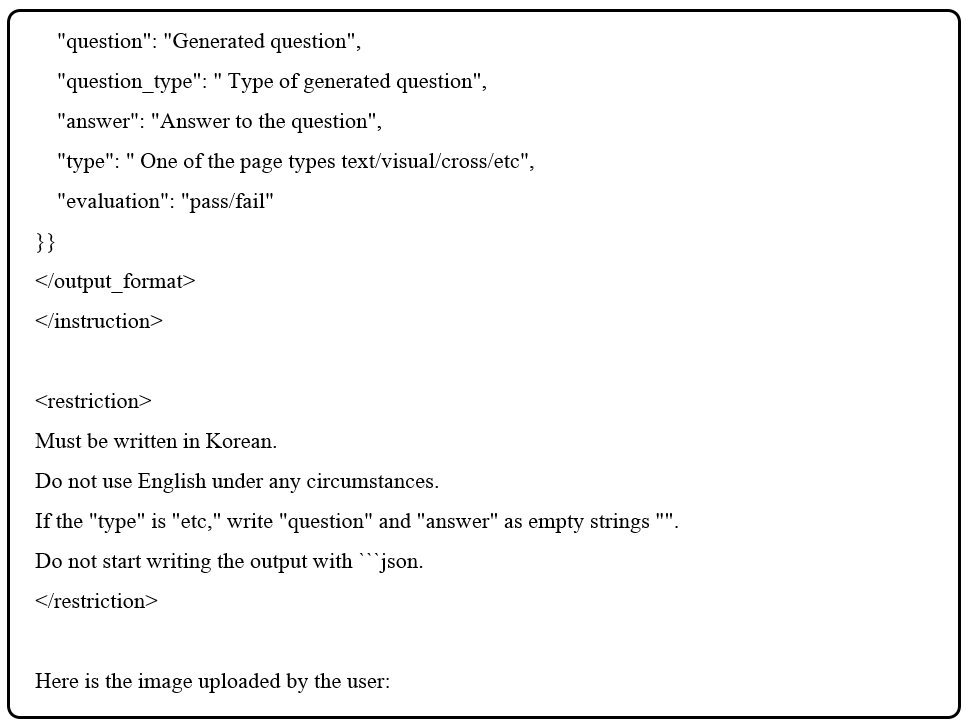}
    \caption{QA generation prompt 4}
    \label{fig:qa_gen_4}
\end{figure}

Figures \ref{fig:qa_gen_1}–\ref{fig:qa_gen_4} illustrate the instruction-based prompt design described in Section 3.3.1.

\subsection{Quality Validation Prompts}
\begin{figure}[htbp]
    \centering
    \includegraphics[width=1\textwidth]{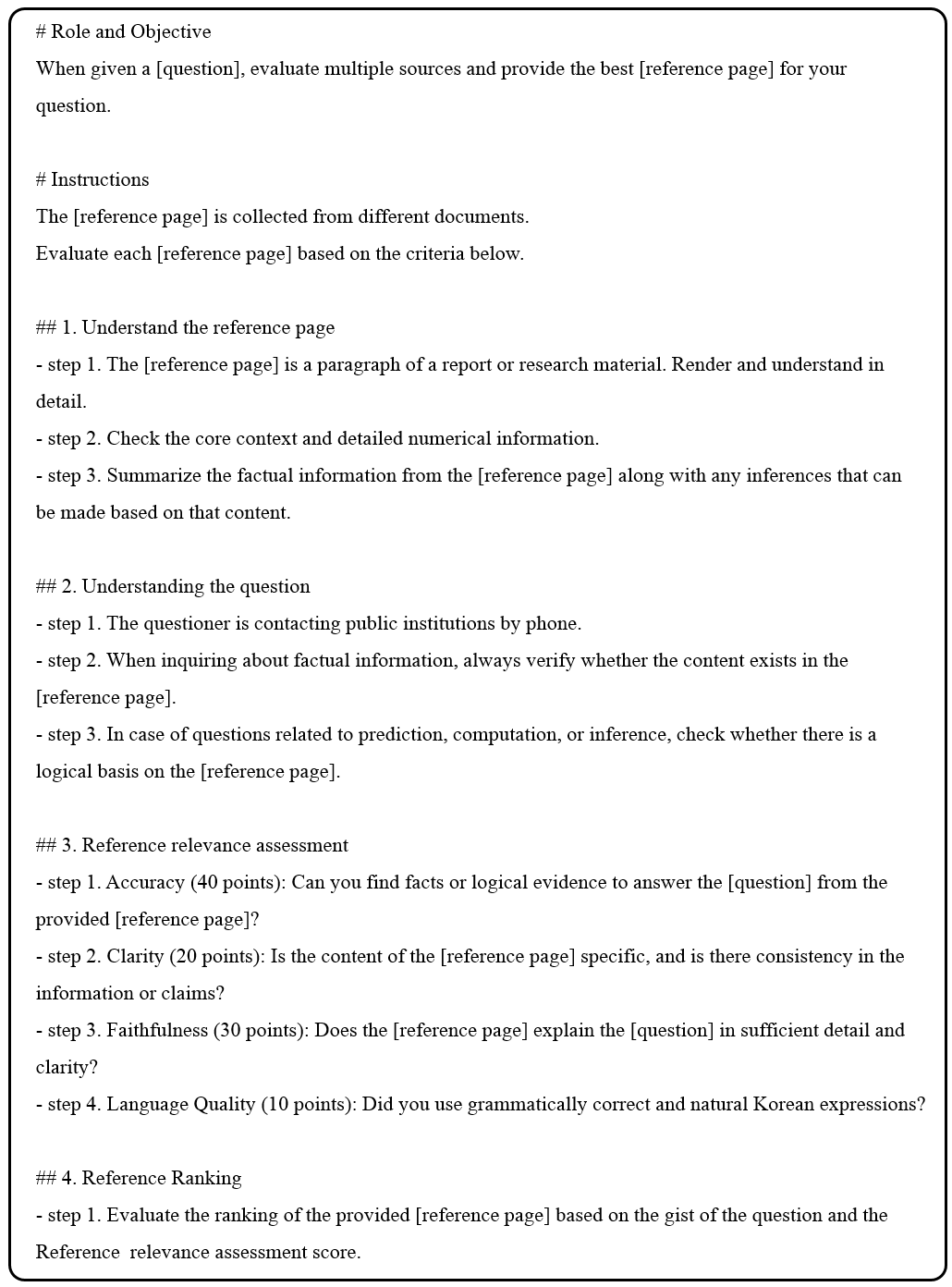}
    \caption{LLM assisted Semantic Verification prompt}
    \label{fig:qa_gpt}
\end{figure}

Figure \ref{fig:qa_gpt} illustrates the LLM-assisted semantic verification prompts described in Section 3.4.2.

\subsection{Human Expert Verification tool}
\begin{figure}[htbp]
    \centering
    \includegraphics[width=1\textwidth]{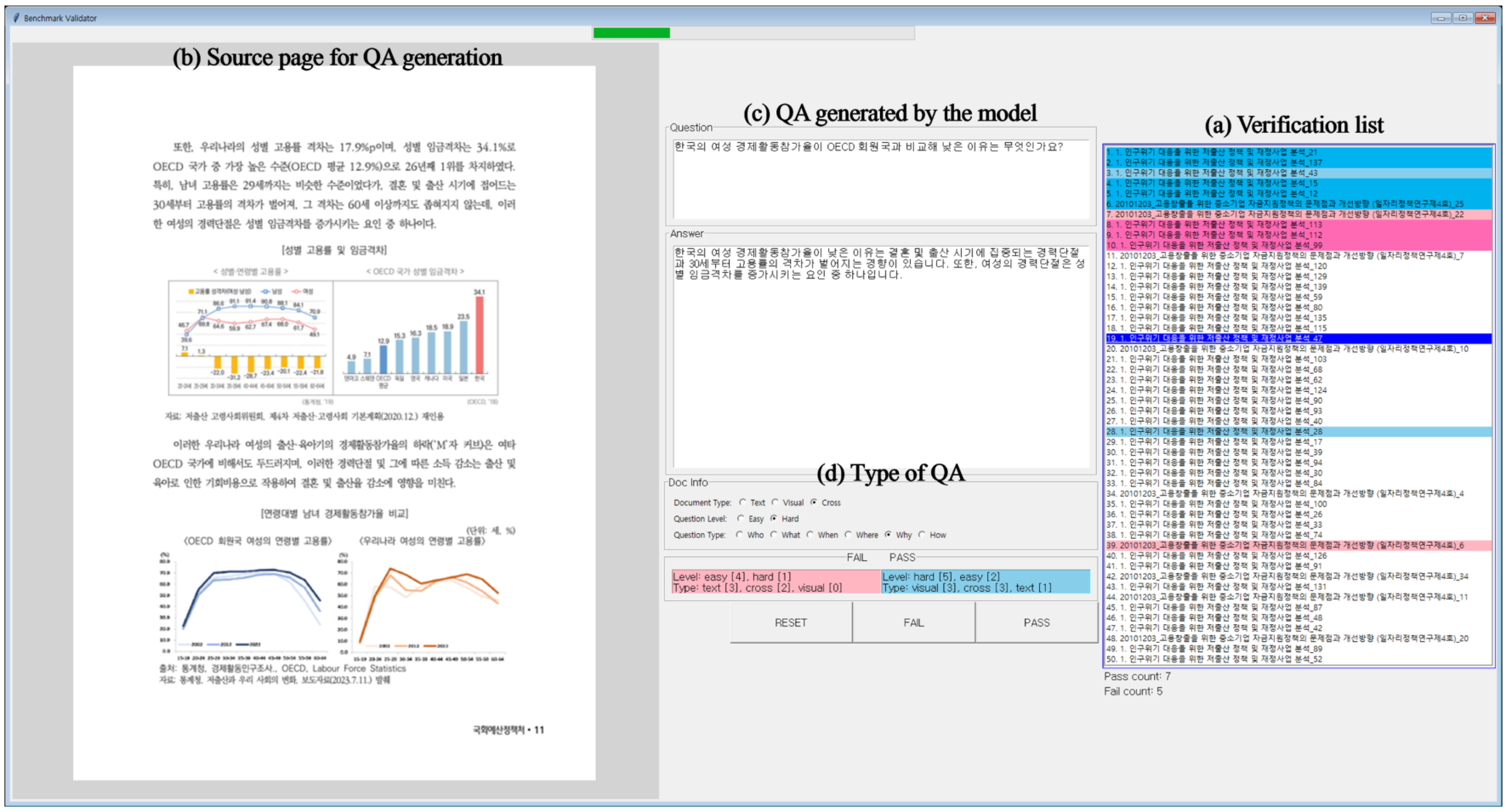}
    \caption{Manual verification tool used in this study:(a) The list of QA pairs to be reviewed is displayed. Each item is color-coded according to the inspection result—Pass (blue) or Fail (red)—and only the failed cases are available for additional review.(b) The page used to generate the selected QA pair from list (a) is shown, allowing reviewers to refer to the original source during inspection.(c) The QA content is displayed in a text editor, where reviewers can directly correct factual inaccuracies or unnatural expressions.(d) The QA type initially classified by the model is shown, and reviewers can modify it by selecting the appropriate option via radio buttons.}
    \label{fig:human_expert}
\end{figure}

Figure \ref{fig:human_expert} presents the manual verification tool used in the study. This figure provides an overview of the interface and workflow employed during the final human review phase described in Section 3.4.3, where researchers conducted full-scale inspection of all QA pairs that passed automated validation.

The tool supports visual cross-checking with the original document pages, direct text correction, and revision of modality labels, ensuring the final benchmark meets both automated and expert-level quality standards.

\end{document}